%% file: main.tex
\documentclass[10pt,twocolumn,letterpaper]{article}

\usepackage[pagenumbers]{cvpr} % To force page numbers, e.g. for an arXiv version

\input{preamble}

\definecolor{cvprblue}{rgb}{0.21,0.49,0.74}
\usepackage[pagebackref,breaklinks,colorlinks,citecolor=cvprblue]{hyperref}
\usepackage[dvipsnames]{xcolor}

\def\paperID{1060} % *** Enter the Paper ID here
\def\confName{CVPR}
\def\confYear{2024}

\title{Towards Realistic Scene Generation with LiDAR Diffusion Models}

\author{
\hspace{0.5cm}Haoxi Ran\\
\hspace{0.5cm}Carnegie Mellon University\\
\hspace{0.5cm}{\tt\small ranhaoxi@cmu.edu}
\and
\hspace{-0.5cm}Vitor Guizilini\\
\hspace{-0.5cm}Toyota Research Institute\\
\hspace{-0.5cm}{\tt\small vitor.guizilini@tri.global}
\and
\hspace{-0.4cm}Yue Wang\\
\hspace{-0.4cm}University of Southern California\\
\hspace{-0.4cm}{\tt\small yue.w@usc.edu}
\and
\url{https://lidar-diffusion.github.io}
}

\begin{document}

\twocolumn[{
\maketitle
\teaser
}]

\input{section/abs.tex}

\input{section/intro.tex}

\input{section/related.tex}

\input{section/method.tex}

\input{section/exp.tex}

\input{section/conclusion.tex}

% \clearpage

{
    \small
    \bibliographystyle{ieeenat_fullname}
    \bibliography{main}
}

\clearpage

% WARNING: do not forget to delete the supplementary pages from your submission 
\input{section/supp/main}

\end{document}

%% file: preamble.tex
\usepackage{subcaption}
\usepackage{graphicx}
\usepackage{amsmath}
\usepackage{amssymb}
\usepackage{booktabs}

\usepackage{enumitem}
\usepackage{algorithm}
\usepackage{listings}
\usepackage{multirow}
\usepackage{pifont}
\usepackage{threeparttable}
\usepackage{makecell}
\usepackage[dvipsnames,table]{xcolor}

\usepackage{url}
\usepackage{adjustbox}
\newcommand\crule[3][black]{\textcolor{#1}{\rule{#2}{#3}}}

\definecolor{baseline}{rgb}{0.85, 0.95, 0.95}
\definecolor{ours}{rgb}{1, 0.93, 0.8}
\definecolor{ours_final}{rgb}{1, 0.82, 0.65}

\definecolor{hp}{rgb}{0.9, 1, 0.85}
\newcommand{\ours}{\cellcolor{ours}}
\newcommand{\hp}{\cellcolor{hp}}

\newcommand{\teaser}{
\vspace{-2em}
\centering
\includegraphics[width=0.95\textwidth,trim=0em 0em 0em 0em,
clip]{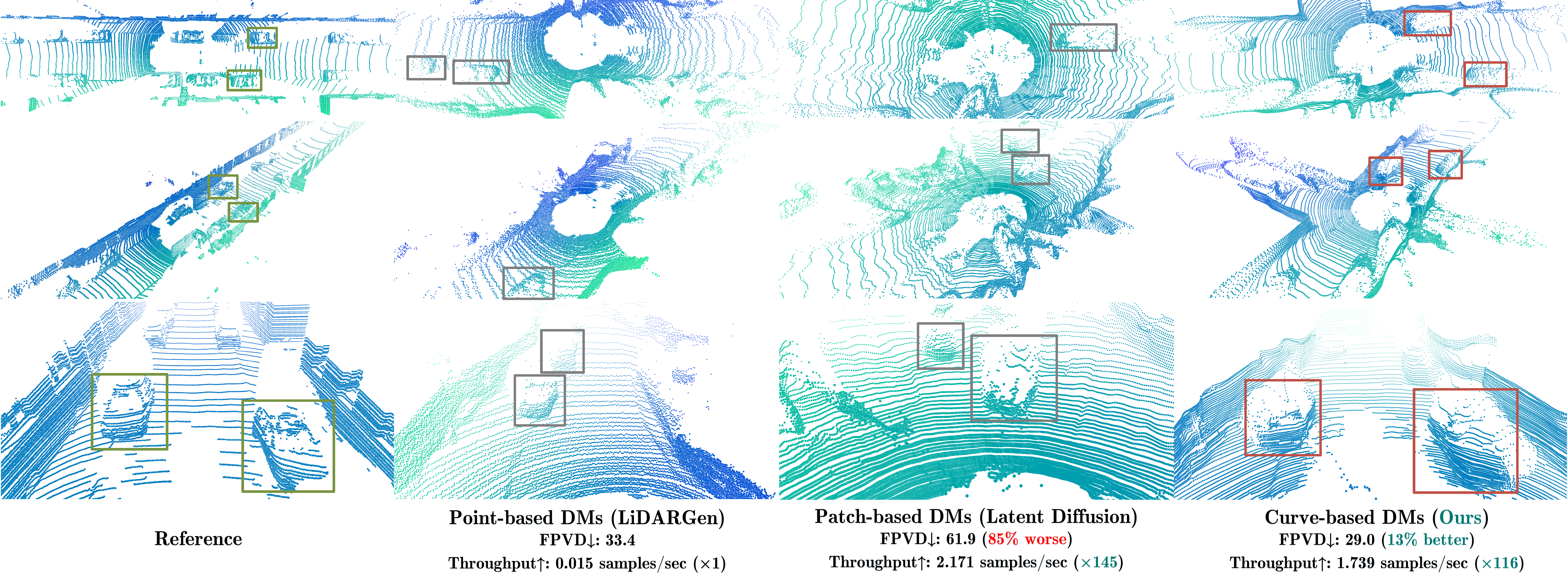}
\vspace{-1.1em}
\captionof{figure}{
Our method (LiDM) establishes a new state-of-the-art in unconditional LiDAR-realistic scene generation, and marks a milestone towards conditional LiDAR scene generation from different input modalities.
}
\label{fig:teaser}
\vspace{1em}
}

%% file: section/abs.tex
\begin{abstract}
\vspace{-1em}
% background
Diffusion models (DMs) excel in photo-realistic image synthesis, but their adaptation to LiDAR scene generation poses a substantial hurdle. This is primarily because DMs operating in the point space struggle to preserve the curve-like patterns and 3D geometry of LiDAR scenes, which consumes much of their representation power. 
% overview
In this paper, we propose \textbf{LiDAR Diffusion Models} (LiDMs) to generate LiDAR-realistic scenes from a latent space tailored to capture the realism of LiDAR scenes by incorporating geometric priors into the learning pipeline. 
% method
Our method targets three major desiderata: pattern realism, geometry realism, and object realism. Specifically, we introduce curve-wise compression to simulate real-world LiDAR patterns, point-wise coordinate supervision to learn scene geometry, and patch-wise encoding for a full 3D object context. 
% experiments
With these three core designs, we establish a new state of the art on unconditional LiDAR generation in 64-beam scenario, while maintaining high efficiency compared to point-based DMs (up to 107$\times$ faster). Furthermore, by compressing LiDAR scenes into a latent space, we enable the controllability of DMs with various conditions such as semantic maps, camera views, and text prompts. 
\end{abstract}

%% file: section/intro.tex
%%%%%%%%% Introduction
\section{Introduction}

%%%%%%%%%% Background of Diffuision models %%%%%%%%%%
Recent years have observed a surge of conditional generative models that are capable of generating visually appealing and highly realistic images. Among them, diffusion models (DMs) have emerged as one of the most popular methods, thanks to its unexceptionable performance. To enable generation with arbitrary conditions, Latent Diffusion Models (LDMs)~\citep{rombach2022high} combine the cross-attention mechanism with a convolutional autoencoder to generate high-resolution images. Its subsequent extensions (\eg, Stable Diffusion~\cite{stablediffusion}, Midjourney~\cite{midjourney}, ControlNet~\citep{zhang2023adding}) further boosted its potential for conditional image synthesis.

%%%%%%%%%% Transition from Image Diffusion to LiDAR Diffusion %%%%%%%%%%
% transition
This success leads us to inquire: can we apply controllable DMs to LiDAR scene generation for autonomous driving and robotics? 
% layout2lidar
For instance, given a collection of bounding boxes, can these models synthesize corresponding LiDAR scenes, thus turning these bounding boxes into high-quality and expensive labeled data? 
% camera2lidar
Alternatively, is it possible to generate a 3D scene solely from a set of images?
% text2lidar
Or even more ambitiously, can we design a language-driven LiDAR generator for controllable simulation?
% conclusion: design conditioning-enabled LiDMs
To answer these interleaved questions, our goal is to design DMs that incorporate a diverse set of conditions (\eg, layouts, camera views, text) to generate LiDAR-realistic scenes.

%%%%%%%%%% Background of Diffuision models for LiDAR Generation %%%%%%%%%%
% transition
To that end, we glean insights from recent works of DMs for autonomous driving. 
% introduce LiDARGen
In \cite{zyrianov2022learning}, \textit{point-based} DM (\ie, LiDARGen) is introduced to unconditional LiDAR scene generation. 
% shortcoming: fail to preserve patterns
However, this model tends to produce noisy backgrounds (\eg, roads, walls) and ambiguous objects (\eg, cars), leading to a failure in generating LiDAR-realistic scenes (\cf Fig.~\ref{fig:teaser}). 
% shortcoming: efficiency
In addition, applying diffusion on points without any compression can computationally slow down the inference process.
% introduce Latent Diffusion
Moreover, the direct application of \textit{patch-based} DMs (\ie, Latent Diffusion~\cite{rombach2022high}) to LiDAR scene generation yields unsatisfactory performance both qualitatively and quantitatively (\cf Fig.~\ref{fig:teaser}). 

%%%%%%%%%% Overview of LiDAR Diffusion %%%%%%%%%%
% overview
To enable conditional LiDAR-realistic scene generation, we thereby propose a \textit{curve-based} generator, termed \emph{LiDAR Diffusion Models} (LiDMs), to answer the aforementioned questions and tackle the shortcomings of recent works. LiDMs are capable of processing arbitrary conditions, such as bounding boxes, camera images, and semantic maps. 
% background of LiDAR representation
LiDMs leverage range images as the representations of LiDAR scenes, which are prevalent in various downstream tasks such as detection~\cite{li2016vehicle, meyer2019lasernet}, semantic segmentation~\cite{wu2018squeezeseg, milioto2019rangenet++}, and generation~\cite{zyrianov2022learning}.
% choice of generating target
This choice is grounded in the reversible and lossless conversion between range images and point clouds, along with the substantial benefits from the highly optimized 2D convolutional operation.
To grasp the semantic and conceptual essence of LiDAR scenes during the diffusion process, our approach leverages encoded points of LiDAR scenes into a perceptually equivalent latent space before the diffusion process.

% towards liDAR-realistic generation
To further improve the realistic simulation of real-world LiDAR data, we focus on three key components: pattern realism, geometry realism, and object realism. 
% pattern realism
 First, We leverage curve-wise compression in the auto-encoding process to maintain the curve patterns of points, motivated by \cite{stearns2023curvecloudnet}.
% geometry realism
Second, to achieve geometry realism, we introduce point-wise coordinate supervision to imbue our auto-encoder with the understanding of scene-level geometry.
% object realism
Lastly, we enlarge the receptive field by incorporating an additional patch-wise down-sampling strategy to capture the full context of visually large objects.
% conclusion
Augmented by these proposed modules, the resulting perceptual space enables DMs to efficiently synthesize high-quality LiDAR scenes (\cf Fig.~\ref{fig:teaser}), while also exhibiting superior performance with a $\times$107 speedup compared to point-based DMs (assessed on one NVIDIA RTX $3090$), and supporting arbitrary types of image-based and token-based conditions.
%%%%%%%%%% Experiments and Evaluation %%%%%%%%%%
We summarize our key contributions as follows:
\begin{itemize}\setlength\itemsep{0em}
\item We propose a novel LiDAR Diffusion Model (LiDM), a generative model that consumes arbitrary input conditions for LiDAR-realistic scene generation. To the best of our knowledge, this is the first method capable of LiDAR scene generation from multi-modal conditions.
\item We introduce \textit{curve-wise compression} to maintain realistic LiDAR patterns, \textit{point-wise coordinate supervision} to regularize models for scene-level geometry, and \textit{patch-wise encoding} to fully capture the context of 3D objects.
\item We introduce three metrics for thorough and quantitative evaluation of the quality of generated LiDAR scenes in the perceptual space, comparing representations including range images, sparse volumes, and point clouds.
\item Our method achieves \textit{state-of-the-art} on unconditional scene synthesis under 64-beam scenario, while realizing a $\times$107 speedup compared to point-based DMs. 
\end{itemize}

%% file: section/related.tex
\section{Related Work}

\paragraph{Diffusion Models.} Diffusion models (DMs)~\cite{sohl2015deep} shows great success in image synthesis~\cite{dhariwal2021diffusion, kingma2021variational, ho2020denoising} in pixel space.
Instead of applying diffusion to pixel space, Latent Diffusion Models (LDMs)~\cite{rombach2022high} adopt a perceptually equivalent latent space for DMs. Their larger-scale applications of LDMs (\eg, Stable Diffusion~\cite{stablediffusion}, Midjourney~\cite{midjourney} further boost the community of DMs.
Recent applications of DMs, including language-guided DMs (\eg, Glide~\cite{nichol2021glide}, DALL-E2~\cite{ramesh2022hierarchical}) and other controllable DMs (\eg, ControlNet~\cite{zhang2023adding}), also reveal the great potential of DMs.

\paragraph{3D Diffusion Models.} 3D diffusion models represent a crucial branch of DMs, offering the capability to generate high-quality samples across various 3D modalities. This includes point clouds~\cite{Zhou_2021_ICCV, Luo_2021_CVPR, melaskyriazi2023projection, tyszkiewicz2023gecco}, meshes~\cite{gupta20233dgen, Liu2023MeshDiffusion, lyu2023controllable}, and implicit fields~\cite{shue20223d, zeng2022lion, li2022diffusion, cheng2022sdfusion, hu2023neural, li20233dqd, erkoç2023hyperdiffusion}. 
Recently, Point-E~\cite{nichol2022point}, a language-guided DM, has demonstrated efficient capabilities in generating high-quality hand-crafted 3D models based on a large-scale 3D dataset.

\paragraph{LiDAR Scene Generation.} Given the larger-scale and complex scenes, treating LiDAR point clouds as a point cloud generation task akin to hand-crafted 3D models encounters difficulties.
In \cite{caccia2019deep}, the authors explore the possibilities of generative models in LiDAR scenes by providing two solutions, LiDARVAE and LiDARGAN. 
\cite{xiong2023learning} and \cite{zyrianov2022learning} further introduce vector-quantized variational autoencoder and point-based diffusion model, respectively, to generate satisfactory samples of LiDAR scenes.
However, these aforementioned methods still fail to generate LiDAR-realistic scenes as they may overlook the curve-like structures and geometric details inherent in LiDAR data.

\paragraph{LiDAR Scene Simulation.} LiDAR simulation produces LiDAR point clouds through physics-based simulators~\cite{dosovitskiy2017carla, koenig2004design} or data-driven simulators~\cite{manivasagam2020lidarsim, hu2021pattern, amini2022vista, wang2022learning}. 
Physics-based LiDAR simulators (\eg, CARLA~\cite{dosovitskiy2017carla}) use raycasting to project rays from the sensor's origin onto the environment's geometry to simulate LiDAR by calculating intersections.
Benefiting from producing realistic LiDAR scenes, data-driven LiDAR simulation gains great attention in the community. A pioneering work LiDARSim~\cite{manivasagam2020lidarsim} adopts deep learning models to produce deviations from physics-based simulations to generate realistic LiDAR point clouds.

%% file: section/method.tex
\section{LiDAR Diffusion Models}

\begin{figure}
\begin{center}
\includegraphics[width=0.48\textwidth]{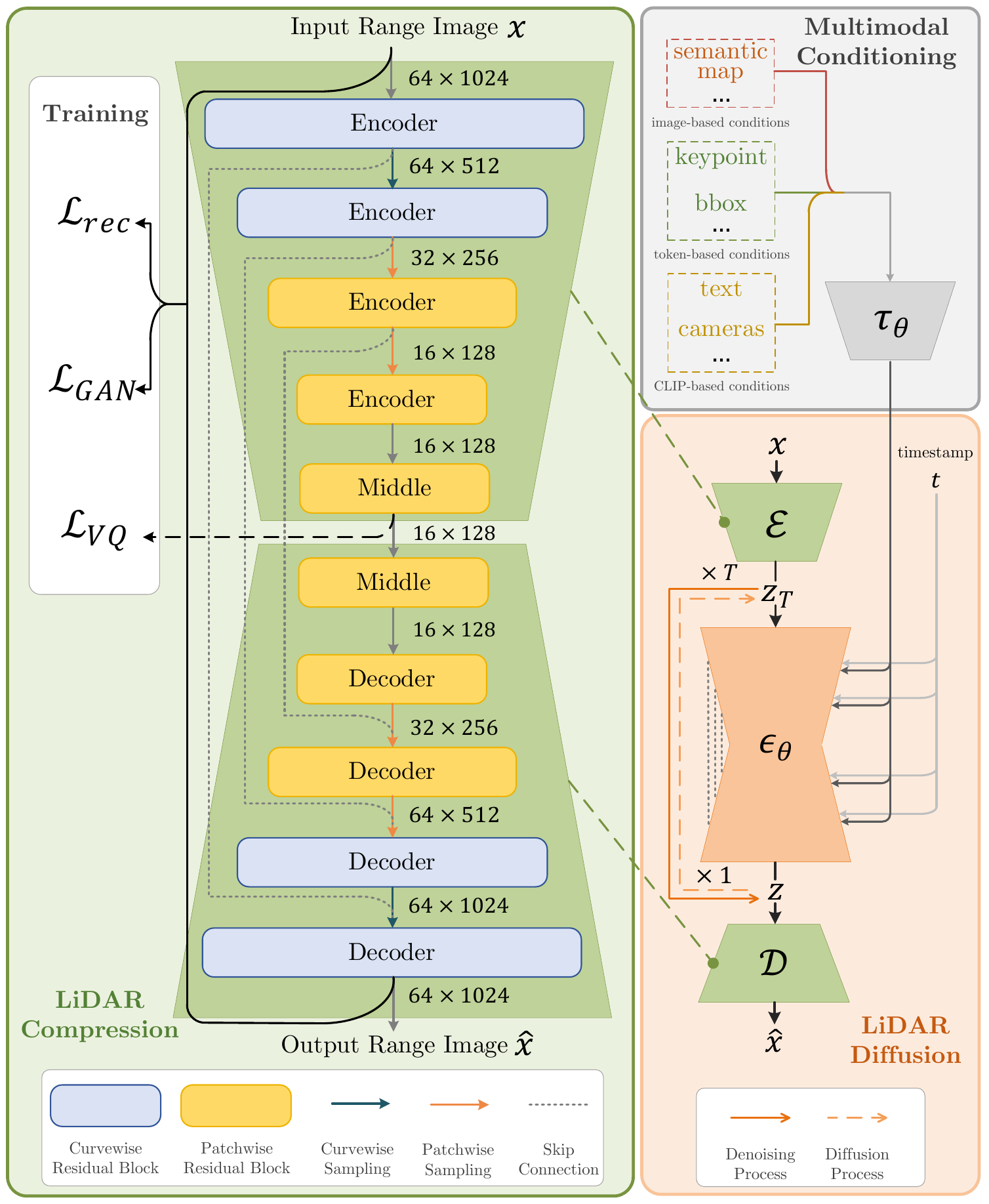}
\end{center}
% \vspace{-0.1in}
   \caption{An overview of LiDMs on 64-beam data, which includes three parts: LiDAR compression (\cf Sec.~\ref{sec:real} \& \ref{sec:train}), Multimodal Conditioning (\cf Sec.~\ref{sec:condition}), and LiDAR Diffusion (\cf Sec.~\ref{sec:train}).}
\label{fig:overview} 
% \vspace{-0.2in}
\end{figure}

In this section, we present \textbf{Li}DAR \textbf{D}iffusion \textbf{M}odels (LiDMs) with details. An overview is shown in Fig.~\ref{fig:overview}.
In Sec.~\ref{sec:rep}, we discuss our design choice of range images for data representation. 
In Sec.~\ref{sec:problem}, we formulate the task of LiDAR scene generation with LiDMs.
In Sec.~\ref{sec:real}, we explore the design of LiDAR compression in terms of both pattern and scene geometry realism. 
In Sec.~\ref{sec:condition}, we present potential applications of LiDM based on multi-modal conditions.
In Sec.~\ref{sec:train}, we define the training objectives in the stages of LiDAR compression and LiDAR Diffusion.
Finally, in Sec.~\ref{sec:eval}, we describe three novel perceptual metrics designed to quantitatively analyze the sample quality for LiDAR scene generation.

\subsection{Data Representation}
\label{sec:rep}

LiDAR data can be represented by different modalities. 
Since our goal is to simulate raw LiDAR output, we have to choose a representation that is losslessly converted to and from raw data, which eliminates optional voxelization. 
To this end, we consider two choices, namely 3D point clouds and 2D range images. 
Both modalities have been successfully explored in previous works, albeit in different settings: 
point cloud generation~\cite{yang2019pointflow, yang2018foldingnet, achlioptas2018learning} often focuses on human-crafted objects due to the efficiency limitation,
while range image generation~\cite{zyrianov2022learning, caccia2019deep} is tailored for larger-scale scenes such as LiDAR scenes.
This separation indicates a clear preference towards range images for our settings.

\subsection{Problem Formulation}
\label{sec:problem}

Both the input and the output of our generators are represented as range images, denoted as $x$ and $\hat{x}$, respectively, where $x, \hat{x}\in\mathbb{R}^{H \times W}$, and $H$ and $W$ are the height and width of the range image, respectively. 
For one pixel, defined by its normalized 2D location and range value, we can directly compute its depth, yaw, and pitch given pre-defined parameters. 
Specifically, we define the 3D coordinate $p=[\alpha,\beta,\gamma]$ of the pixel with:
\begin{align}
&\alpha=cos(\mathrm{yaw}) \times cos(\mathrm{pitch}) \times \mathrm{depth}, \\
&\beta=-sin(\mathrm{yaw}) \times cos(\mathrm{pitch}) \times \mathrm{depth}, \\
&\gamma=sin(\mathrm{pitch}) \times \mathrm{depth}.
\end{align}

Our LiDMs pipeline can be split into two parts: LiDAR compression and diffusion process. 
For LiDAR compression, we use an encoder $\mathcal{E}$ to compress range images into latent code $z=\mathcal{E}(x)$, where $z\in\mathbb{R}^{h \times w \times d}$, and a decoder $\mathcal{D}$ to decode it into $\hat{x}=\mathcal{D}(z)=\mathcal{D}(\mathcal{E}(x))$.
For the diffusion process, following standard practice~\cite{rombach2022high}, we denote an equally weighted sequence of unconditional denoising auto-encoders as $\epsilon_\theta(z_t, t)$, with timestamp $t=1 \dots T$.

\subsection{Towards LiDAR-Realistic Generation}
\label{sec:real}

\noindent\textbf{Pattern Realism}~~
The presence of curves is a common pattern in LiDAR scenes. 
A curve $c_i=[p_1,...,p_{n_i}]$ is represented by a sequence of $n_i$ points where consecutive pairs are connected by a polyline. Furthermore, as defined in \cite{stearns2023curvecloudnet}, a LiDAR point cloud is equivalent to a set of curves, namely a curve cloud $C=\{c_1,...,c_m\}$. While we cannot directly apply the concept of curve cloud to range images, we can still explore curve-like structures. 

Given that each beam of a LiDAR sensor sequentially captures 3D points by scanning the scene horizontally, we can safely assume that each curve is stored in only one row of a range image, and is represented as a continuous segment of pixels. 
Thus, we design our auto-encoders through curve-wise compression, which results in \textit{horizontally} downsampled range images, by a factor $f_c:=2^\eta$, where $\eta\in\mathbb{N}$. 
To achieve curve-wise compression, we implement with these details: 1) the kernel size of convolutions inside each curve-wise residual block is $1\times4$, instead of the traditional $3\times3$ used on images; 2) each downsampling or upsampling layer is applied only horizontally; 3) the padding is circular considering the two sides of an range image to be connected end to end. 
Through this implementation, we effectively preserve curve-like structures in a perceptually equivalent space.

\noindent\textbf{Geometry Realism}~~
Preserving scene-level geometry is another key aspect in LiDAR-realistic generation. To achieve this, our model should possess the capability to clearly distinguish between objects and the background, demonstrating sensitivity to the contours of 3D geometry.
However, the conversion of point clouds to range images (as described in Sec.~\ref{sec:problem}) may lead to a loss of geometry.
Thus, we introduce a novel point-wise coordinate supervision to enhance the understanding of autoencoders in 3D space.
Point clouds are informative to describe the geometry of LiDAR scenes through coordinates, but due to irregularity, we cannot directly apply point cloud distance loss functions (\eg, Chamfer Distance~\cite{fan2017point}) to the training of autoencoders.
For this purpose, we design a simple manner by supervising the converted coordinate of each pixel between the input and output range images. Note that, the converted coordinate-based images are in the shape of $H \times W \times 3$. 
Point-wise coordinate supervision includes both a pixelwise 3D distance loss and an adversarial objective on the coordinate-based images.

\noindent\textbf{Object Realism}~~
Achieving object realism is challenging, but an important aspect of recovering reasonable and complete shapes. 
While curve compression effectively captures patterns in LiDAR scenes, it suffers from a restricted receptive field when capturing the complete context of 3D objects, particularly for visually larger objects in range images (\ie, objects near the ego-center).
Thus, we introduce patch-wise blocks in the intermediate layers of autoencoders for patch-wise encoding, which is qualitatively effective as a way to improve the synthesis quality of objects. We define the factor of downsampling during patch-wise encoding as $f_p=2^\mu$, $\mu\in\mathbb{N}$, and thus $h=H/f_p$, $w=W/(f_c \times f_p)$.
%
%
%
%
%
%

%%%%%%%%%%%%%%%%%%%%%%%%%%%%%%%%%%%%%%%%%%%%%%%%%%%%%%%%%%%%%%

%%%%%%%%%%%%%%%%%%%%%%%%%%%%%%%%%%%%%%%%%%%%%%%%%%%%%%%%%%%%%%

\subsection{Multimodal Conditioning}
\label{sec:condition}
Previous works~\cite{rombach2022high, zhang2023adding} have shown DMs' capability to model conditional distributions.
Utilizing our LiDMs, we further introduce multimodal conditioning to realize the significant potential for downstream tasks within the domain of autonomous driving. 
Typically, two types of conditions can serve as inputs in LiDAR scenes: image-based conditions (\eg, semantic maps), and token-based conditions (\eg, bounding boxes, keypoints).
We approach the applications of image-based conditioning as image-to-image translation tasks~\cite{isola2017image}, while we employ the cross-attention mechanism to handle token-based conditions, aligning with a widely adopted practice~\cite{rombach2022high}.
To broaden the scope of conditional LiDAR generation, we introduce \textbf{Camera-to-LiDAR} task by extracting the global features of each view using a pretrained latent space provided by CLIP~\cite{radford2021learning}. 
Due to the spatial mismatch between multiple camera views and a LiDAR point cloud, we cannot directly treat Camera-to-LiDAR generation as another image-to-image translation task.
Therefore, for a LiDAR scene, in contrast to tasks employing the entire image-based condition, we guide LiDMs with a condition representation formed by concatenating global features of all camera views.

Most recently, the paradigm of contrastive image-text pretraining (\eg, CLIP~\cite{radford2021learning}) has demonstrated remarkable progress in zero-shot learning~\cite{ramesh2021zero}. CLIP enables zero-shot understanding across diverse generation tasks of text-to-image~\cite{ramesh2022hierarchical}, text-to-video~\cite{xu2021videoclip}, \etc. 
To explore the potential applications of language-guided autonomous driving~\cite{jain2023ground} for LiDAR generation, we leverage the text-image latent space of CLIP to encode descriptive prompts for LiDMs. Consequently, we can seamlessly transition the Camera-to-LiDAR task into a novel \textbf{Text-to-LiDAR} generation.

\subsection{Training Objectives}
\label{sec:train}

\noindent\textbf{LiDAR Compression}~~
Through curve-wise compression, point-wise coordinate supervision and patch-wise encoding, we design autoencoders to compress range images. 
To train these autoencoders, we adopt a set of objectives, including a pixelwise $L_1$ reconstruction objective $L_{rec}$, a curve-based adversarial objective adapted from \cite{isola2017image} $L_\mathrm{GAN}$ and a vector quantization regularization~\cite{van2017neural} $L_\mathrm{VQ}$. 
Specifically, we compute both $L_{rec}$ and $L_\mathrm{GAN}$ with both range and coordinate-based images as input. 
We compute $L_{rec}$ in the parts of pixelwise $L_1$ loss with range images and pixelwise coordinate distance loss with coordinate-based images:
\begin{equation}
\mathcal{L}_{rec}\left(x\right)=\mathbb{E}_x[\|x-\hat{x}\| + \lambda\|p-\hat{p}\|_2^2],
\end{equation}
where $\lambda$ is a scale factor for the supervision of coordinate-based images.
Additionally, we compute $L_{GAN}$ by feeding both range images and coordinate-based images into our CurveGAN, a variant of PatchGAN~\cite{isola2017image} by applying curve-wise compression in the first stage. 
We define the adversarial objective as:
\begin{equation}
\mathcal{L}_{\mathrm{GAN}}(x)=\mathbb{E}_x[\log \mathcal{D}([x, p])+\log (1-\mathcal{D}([\hat{x}, \hat{p}]))],
\end{equation}
where $[\dots]$ means concatenation operation. Besides, we utilize the loss of vector quantization~\cite{van2017neural} $L_\mathrm{VQ}$ to learn a codebook of range image constituents. We incorporate the vector quantization layer in the decoder, following the implementation of \cite{esser2021taming}. Overall, the complete training objective of autoencoders is:
\begin{equation}
L_\mathrm{AE} = \mathcal{L}_{rec} + \mathcal{L}_{\mathrm{GAN}} + \mathcal{L}_\mathrm{VQ}.
\end{equation}

\noindent\textbf{LiDAR Diffusion}~~
As probabilistic models, DMs~\cite{sohl2015deep} aim to comprehend a data distribution through a progressive denoising process on variables sampled from a Gaussian distribution. In image synthesis, previous works have employed DMs either in the pixel space~\cite{dhariwal2021diffusion, ho2020denoising} or a latent space~\cite{rombach2022high}.
In this paper, we enable DMs to leverage a low-dimensional latent space created by our autoencoders. This space adequately preserves LiDAR patterns and the geometry of scenes and objects while maintaining computational efficiency. Furthermore, it empowers DMs to focus on the semantic information of large-scale LiDAR scenes.
We define the objective of our \textit{unconditional} LiDMs:
\begin{equation}
L_\mathrm{LiDM}=\mathbb{E}_{\mathcal{E}(x), \epsilon \sim \mathcal{N}(0,1), t}\left[\left\|\epsilon-\epsilon_\theta\left(z_t, t\right)\right\|_2^2\right],
\end{equation}
where $\epsilon_\theta\left(z_t, t\right)$ is a UNet~\cite{ronneberger2015u} backbone with timestamp conditioning. For inference, we obtain sampled range images by decoding the denoised latent code $z$ with $\mathcal{D}$. Similarly, given an input condition $y$, we define the objective of our \textit{conditional} LiDMs as:
\begin{equation}
L_\mathrm{cLiDM}=\mathbb{E}_{\mathcal{E}(x), y, \epsilon \sim \mathcal{N}(0,1), t}\left[\left\|\epsilon-\epsilon_\theta\left(z_t, t, \tau_\theta(y)\right)\right\|_2^2\right],
\end{equation}
where $\tau_\theta$ is a condition encoder responsible for converting $y$ into an accessible representation for the UNet backbone.

\subsection{Evaluation Metrics}
\label{sec:eval}
To evaluate LiDAR generative models, previous works on LiDAR scene generation~\cite{caccia2019deep, zyrianov2022learning} commonly utilize statistical metrics as proposed in \cite{achlioptas2018learning}. Nonetheless, they may encounter difficulties to quantitatively measure the synthesis quality at a perceptual level. Thus, we design several perceptual level metrics for LiDAR generative models.

Perceptual evaluation (\eg, FID~\cite{heusel2017gans}) stands as a prevalent manner to measure image synthesis quality across a spectrum of  popular image-based generators~\cite{karras2017progressive, karras2019style, karras2020analyzing, ho2020denoising, rombach2022high, dhariwal2021diffusion, saharia2022photorealistic}. Typically, these prior works compute the Fréchet distance~\cite{frechet1957distance} between the data distributions of of real data and generated samples within a perceptual space defined by a pretrained classifier (\eg, Inception model~\cite{szegedy2016rethinking}). 
To enhance comprehension of the performance of LiDAR generative models at a perceptual level, we augment the perceptual evaluation with three Fréchet-distance-based perceptual metrics: Fréchet Range Image Distance (FRID), Fréchet Sparse Volume Distance (FSVD), and Fréchet Point-based Volume Distance (FPVD). 

In the absence of a pretrained classifier specifically tailored for LiDAR scenes, we adopt segmentation-based pretrained models for our evaluations.
Specifically, we calculate FRID, FSVD, FPVD through three simple methods, which include RangeNet++~\cite{milioto2019rangenet++}, MinkowskiNet~\cite{choy20194d}, and SPVCNN~\cite{tang2020searching}, respectively. 
Different from FID, which is computed relying on the global feature of each input image in the final stage, our proposed metrics are founded on the average of the output features of each LiDAR scene in the intermediate stage. 
With the computed features of samples and the real-world data, we enable the performance comparisons in perceptual space between previous LiDAR generators and our LiDMs. \cf the supplement for details.

%% file: section/exp.tex
%%%%%%%%% Experiments

\section{Experiments}

%------------------------------------------------------------------------

\subsection{Experimental Settings}

We train and evaluate our models in the LiDAR scenarios of 32-beam data from nuScenes~\cite{fong2022panoptic}, gathered around the suburbs in Germany, and 64-beam data from KITTI-360~\cite{liao2022kitti}, collected inside the cities. For the 32-beam scenario, we train autoencoders on the full training dataset, containing 286,816 samples, and validate with 10,921 samples. 
For the 64-beam scenario, we train autoencoders on 63,315 samples of 9 sequences (including training and test set) and evaluate on 1,031 samples of one sequence. 
Different from autoencoders, we train and validate LiDMs on the widely adopted subsets of both datasets, which provide various conditions, including bounding boxes, views of multiple cameras or perspective views. 
Specifically, we adopt SemanticKITTI~\cite{behley2019semantickitti} for Semantic-Map-to-LiDAR task. 

Our LiDMs process range images with dimensions of 32$\times$1024 for 32-beam data and 64$\times$1024 for 64-beam data. The pixel values of the range images are computed through binary logarithm followed by scaling. Training is conducted on eight NVIDIA RTX 3090, each with 24GB of GPU memory, and one of them is utilized for inference.

%%%%%%%%%%%%%%%%%%%%%%%%%%%%%%%%%%%%%%%%%%%%%%%%%%%%%%%%%%%%%%
\begin{figure*}
\begin{center}
\includegraphics[width=\textwidth]{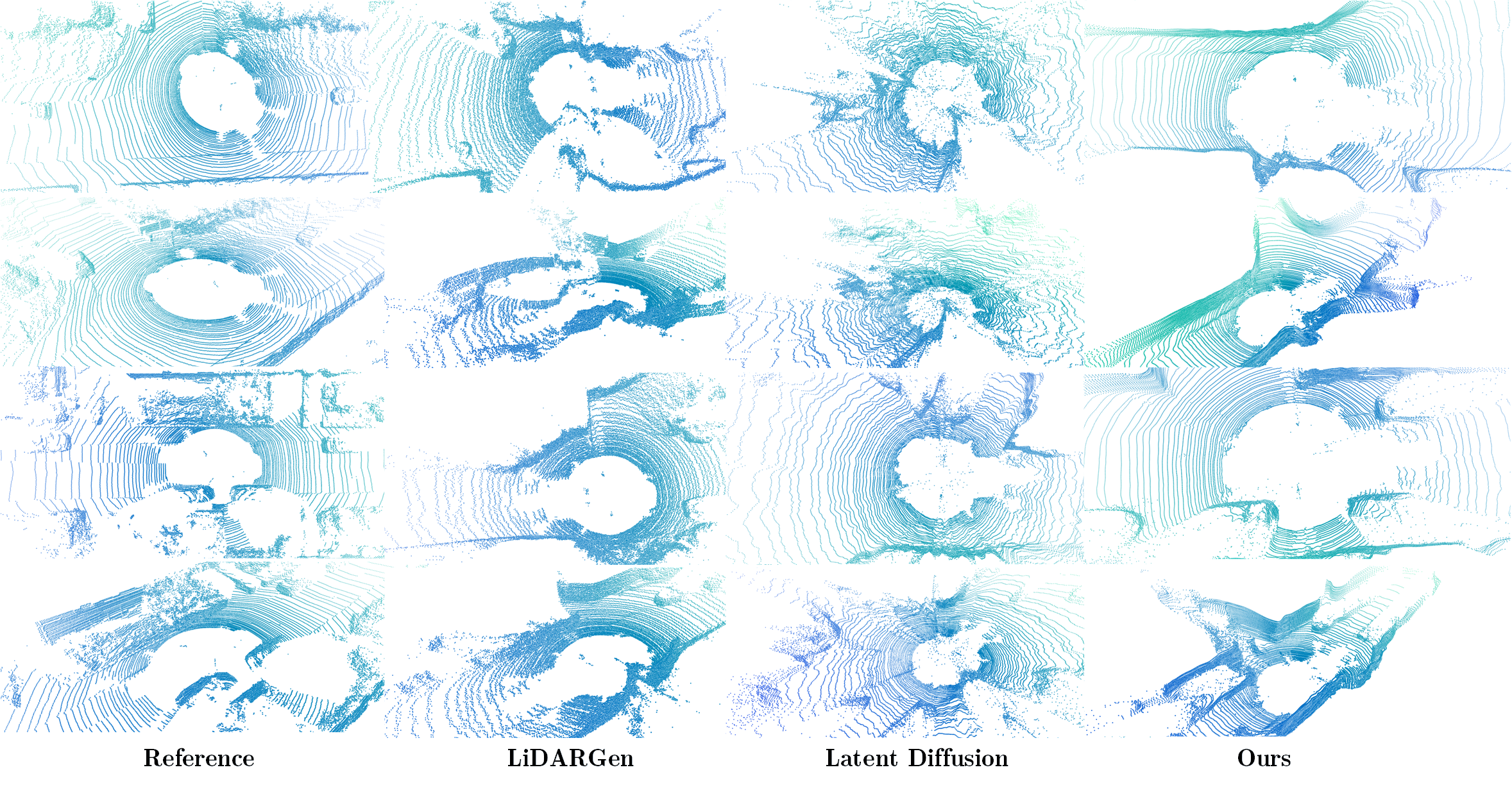}
\end{center}
\vspace{-0.2in}
\caption{Samples from LiDARGen~\cite{zyrianov2022learning}, Latent Diffusion~\cite{rombach2022high}, and our LiDMs on 64-beam scenario.}
\label{fig:uncond} 
\vspace{-3mm}
\end{figure*}

%%%%%%%%%%%%%%%%%%%%%%%%%%%%%%%%%%%%%%%%%%%%%%%%%%%%%%%%%%%%%%

%%%%%%%%%%%%%%%%%%%%%%%%%%%%%%%%%%%%%%%%%%%%%%%%%%%%%%%%%%%%%%
\begin{figure}
\begin{center}
\includegraphics[width=0.48\textwidth]{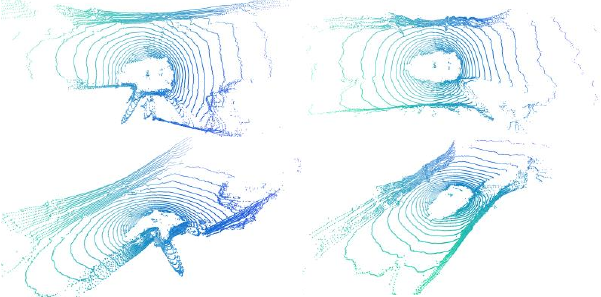}
\end{center}
\vspace{-0.2in}
\caption{Samples from our LiDMs on 32-beam scenario. }
\label{fig:uncond_32} 
\vspace{-3mm}
\end{figure}

%%%%%%%%%%%%%%%%%%%%%%%%%%%%%%%%%%%%%%%%%%%%%%%%%%%%%%%%%%%%%%

\subsection{Unconditional LiDAR Diffusion}

\input{table/uncond_gen}

To verify the effectiveness of our method, we train unconditional LiDMs on 64-beam data from KITTI-360~\cite{liao2022kitti}. Following the practice of FID~\cite{heusel2017gans}, we generate 2000 samples for the evaluation of unconditional generation.
We evaluate the performance of each method using FRID (range image), FSVD (sparse volume), and FPVD (point-based volume), as described in Sec.~\ref{sec:eval}. Together, these three metrics enable the perceptual evaluation to assess the quality of LiDAR scenes generated with different methods.

As shown in Table~\ref{tab:uncond}, with a very limited number of sampling steps (\ie, 50), we establish a new state of the art for almost all considered metrics. Specifically, within 50 sampling steps, our approach outperforms the previous state-of-the-art method (\ie, LiDARGen~\cite{zyrianov2022learning}) by a large margin for all considered metrics. Additionally, with only 50 evaluation steps, our method performs competitive with LiDARGen with a longer diffusion process of 1160 steps. LiDM reports $9.7\%\sim31.0\%$ improvement over baseline model LDM~\cite{rombach2022high} after replacing only the autoencoder. With all the techniques mentioned in Sec.~\ref{sec:real}, LiDM further improves by $10.6\%\sim53.2\%$ over LDM.
For a qualitative comparison, in Fig.~\ref{fig:uncond} we provide examples generated with each model, alongside reference point clouds. We further provide some 32-beam samples for example in Fig.~\ref{fig:uncond_32}.

\subsection{Conditional LiDAR Diffusion}

To further exploit the potential of LiDMs, we implement several variations of conditional LiDAR scene generation, including Semantic-Map-to-LiDAR and Camera-to-LiDAR. For a quantitative analysis, we compare LiDMs to LiDARGen~\cite{zyrianov2022learning} and to our baseline Latent Diffusion~\cite{rombach2022high}, with results reported in Table~\ref{tab:cond}.

\input{table/cond_gen}

%%%%%%%%%%%%%%%%%%%%%%%%%%%%%%%%%%%%%%%%%%%%%%%%%%%%%%%%%%%%%%
\begin{figure}
\begin{center}
\includegraphics[width=0.48\textwidth]{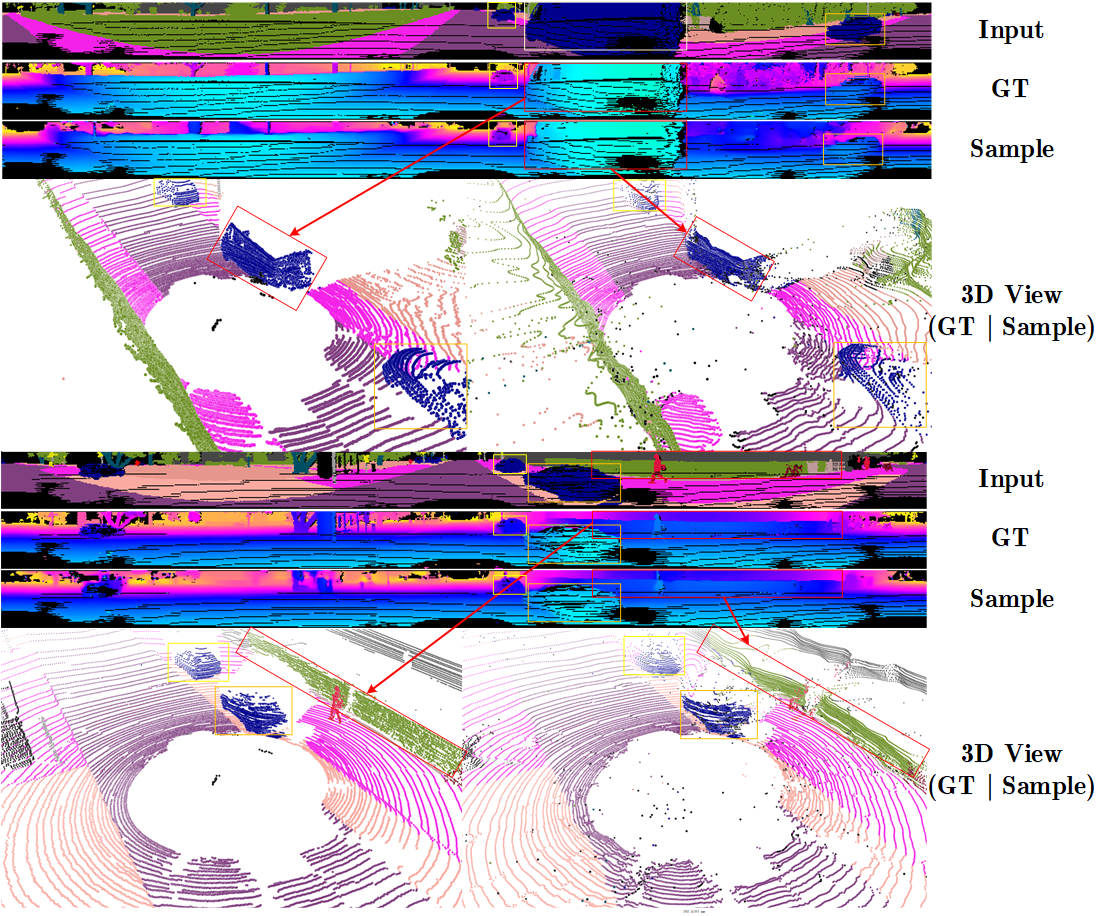}
\end{center}
\vspace{-0.1in}
\caption{Samples from our LiDM for Semantic-Map-to-LiDAR generation on SemanticKITTI~\cite{behley2019semantickitti}.}
\label{fig:map2lidar}
\vspace{-0.2in}
\end{figure}

%%%%%%%%%%%%%%%%%%%%%%%%%%%%%%%%%%%%%%%%%%%%%%%%%%%%%%%%%%%%%%

\noindent\textbf{Semantic-Map-to-LiDAR}~~Transforming semantic maps into RGB images is a typical image-to-image translation task~\cite{isola2017image}. However, it remains underexplored in the context of LiDAR scene generation. 
As shown in our reported results, LiDM outperforms Latent Diffusion and LiDARGen by a substantial margin. Additionally, conditioning LiDMs with semantic maps leads to significant improvements relative to unconditional generation. We argue that having access to such data enhances the understanding of LiDMs at a semantic level, which facilitates the generation of LiDAR-realistic scenes. Fig.~\ref{fig:map2lidar} further illustrates the effectiveness of semantic-map-based conditioning with LiDMs.

%%%%%%%%%%%%%%%%%%%%%%%%%%%%%%%%%%%%%%%%%%%%%%%%%%%%%%%%%%%%%%
\begin{figure}
\begin{center}
\includegraphics[width=0.45\textwidth]{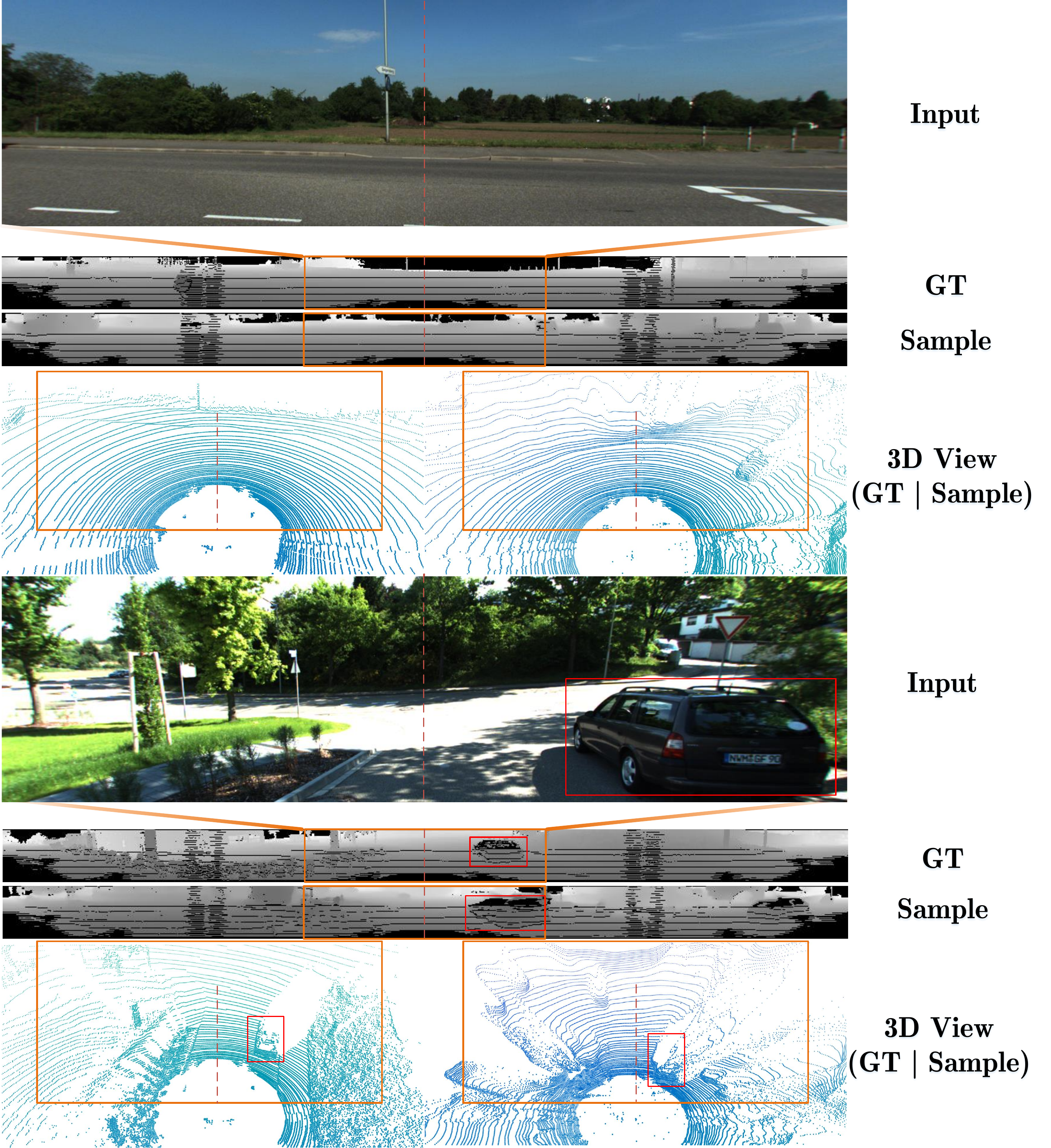}
\end{center}
\vspace{-0.1in}
\caption{LiDM samples for conditional Camera-to-LiDAR generation on KITTI-360~\cite{liao2022kitti}. The orange box indicates the area covered by the input image. For each scene, KITTI-360 provides one perspective, which cover only a part of the scene. Thus, LiDM performs conditional generation for the camera-covered region and unconditional generation for the remaining unobserved regions.}
\label{fig:cam2lidar}
\vspace{-0.2in}
\end{figure}

%%%%%%%%%%%%%%%%%%%%%%%%%%%%%%%%%%%%%%%%%%%%%%%%%%%%%%%%%%%%%%

\noindent\textbf{Camera-to-LiDAR}~~Camera views are commonplace in the context of autonomous driving. To explore the relationship and complementarity between cameras and LiDAR sensor data, we implement Camera-to-LiDAR generation on KITTI-360~\cite{liao2022kitti}. In this setting, LiDMs outperform LiDARGen~\cite{zyrianov2022learning} by over 36\% among all metrics, while also successfully capturing semantic information from camera views. In Fig.~\ref{fig:cam2lidar}, on the top, we see LiDM generating smooth ground from an input image containing a road without any objects. Similarly, on the bottom we see LiDM extracting the semantic information about the presence of a car and generating it on the synthesized LiDAR scene, although it still struggles with its precise scale in 3D space.

%%%%%%%%%%%%%%%%%%%%%%%%%%%%%%%%%%%%%%%%%%%%%%%%%%%%%%%%%%%%%%
\begin{figure}
\begin{center}
\includegraphics[width=0.45\textwidth]{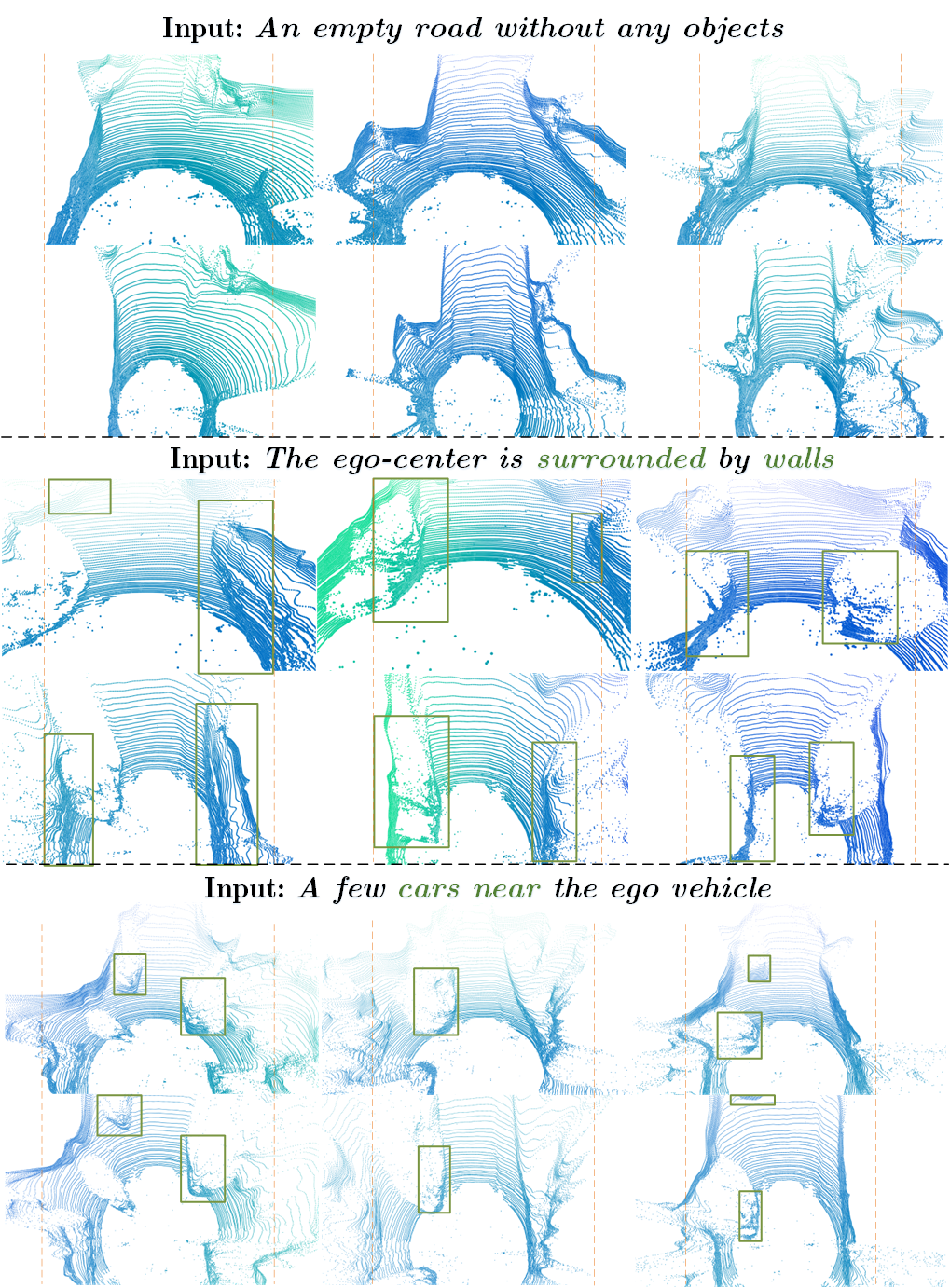}
\end{center}
\vspace{-0.1in}
\caption{LiDM samples for zero-shot Text-to-LiDAR generation on 64-beam scenario. The areas enclosed by orange dotted lines indicate those influenced by the conditioning, and green boxes highlight objects potentially associated with the prompts.}
\label{fig:text2lidar}
\vspace{-0.15in}
\end{figure}

%%%%%%%%%%%%%%%%%%%%%%%%%%%%%%%%%%%%%%%%%%%%%%%%%%%%%%%%%%%%%%

\subsection{Zero-shot Text-to-LiDAR Generation}

Text-to-image learning has become very popular recently due to the introduction of contract language-vision pretraining paradigm~\cite{radford2021learning}. To facilitate LiDAR generation with language-guided conditioning, we introduce zero-shot Text-to-LiDAR generation based on a pretrained Camera-to-LiDAR LiDM, which is transformable to the task of Text-to-LiDAR. Through provided prompts, LiDM can hallucinate possible scenes related to the input prompts. Fig.~\ref{fig:text2lidar} shows some evidence to this argument. However, Text-to-LiDAR LiDM still struggles to generate scenes when presented with complex prompts, primarily due to constraints imposed by the limited amount of available training data.

\subsection{Study on LiDAR-Realistic Generation}

%%%%%%%%%%%%%%%%%%%%%%%%%%%%%%%%%%%%%%%%%%%%%%%%%%%%%%%%%%%%%%
\begin{figure}
\begin{center}
\includegraphics[width=0.23\textwidth]{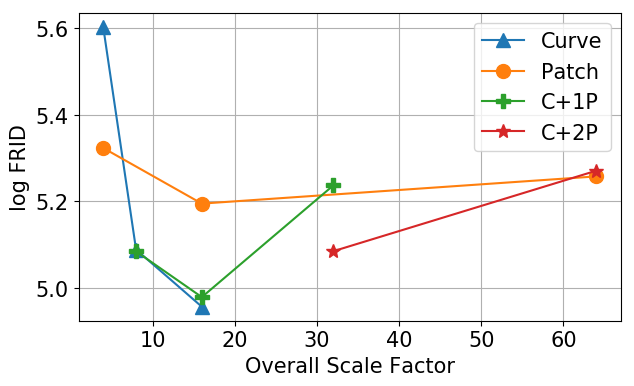}
\includegraphics[width=0.23\textwidth]{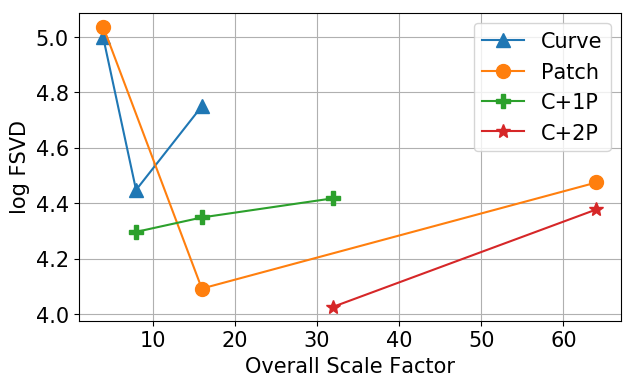}
\end{center}
\vspace{-0.2in}
\caption{Overall scale factor ($f_c \times f_p$) vs sampling quality (FRID \& FSVD). We compare different scales of curve-wise encoding (Curve), patch-wise encoding (Patch), and curve-wise encoding with one (C+1P) or two (C+2P) stages of patch-wise encoding on KITTI-360~\cite{liao2022kitti}.}
\label{fig:ablate} 
\vspace{-0.1in}
\end{figure}

\begin{figure}
\begin{center}
\includegraphics[width=0.48\textwidth]{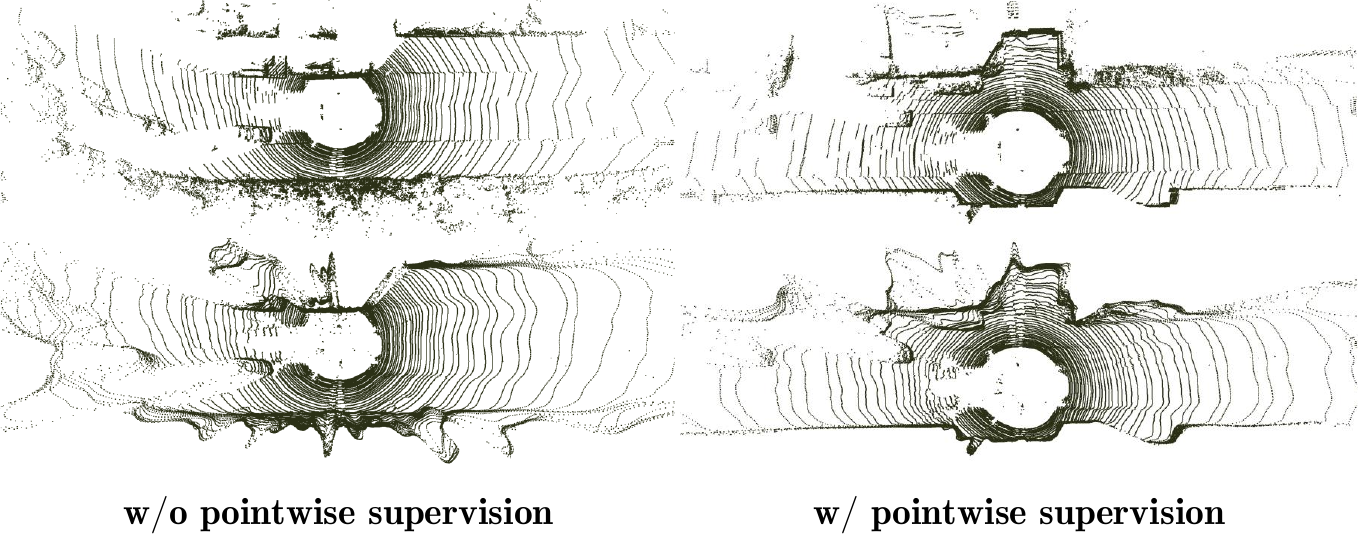}
\end{center}
\vspace{-0.2in}
\caption{Samples from LiDM with or without point-wise supervision, as proposed in Sec.~\ref{sec:real}.}
\label{fig:coord} 
\vspace{-0.1in}
\end{figure}

%%%%%%%%%%%%%%%%%%%%%%%%%%%%%%%%%%%%%%%%%%%%%%%%%%%%%%%%%%%%%%

We explore our designed autoencoders for LiDAR compression and ablate on our proposed point-wise coordinate supervision. 
To analyze the behavior of LiDAR compression in terms of curve-wise and patch-wise encoding, we conduct experiments on various scale factors. 
The results in Fig.~\ref{fig:ablate} show that curve-wise encoding generally performs better than patch-wise encoding. However, by introducing one stage of patch-wise encoding, we allow the autoencoders to further compress range images while maintaining competitive performance. 
To balance between performance and compression rate, we chose $f_c=2$ and $f_p=4$ as the default settings for our experiments. 

Additionally, we study the effectiveness of our proposed point-wise coordinate supervision. 
The visualization in Fig.~\ref{fig:coord} illustrates that point-wise coordinate supervision aids autoencoders in preserving scene-level geometry by reconstructing sharper boundaries in 3D space.

\subsection{Efficiency Analysis}

\input{table/efficiency}

Efficiency is particularly important for LiDAR generative models, specially when considering adoption to downstream tasks. To investigate this aspect of LiDMs, we provide an overhead comparison between LiDMs and the previous state-of-the-art point-based DM, LiDARGen~\cite{zyrianov2022learning}, in terms of throughput (\emph{samples/sec}), inference speed (\emph{steps/sec}). As shown in Table~\ref{tab:efficiency}, the throughput and inference speed of LiDM is around $\times$107 and $\times$4.6 faster than LiDARGen~\cite{zyrianov2022learning}, respectively, which shows the superiority of our method in terms of efficiency.

%% file: table/uncond_gen.tex
\begin{table}[t]
\centering
\begin{footnotesize}
\begin{adjustbox}{max width=\linewidth}
\begin{tabular}{ccccccc}
\toprule
& \multicolumn{3}{c}{\textbf{Perceptual}} & & \multicolumn{2}{c}{\textbf{Statistical}} \\ 
\cmidrule{2-4} \cmidrule{6-7}
Method & FRID $\downarrow$ & FSVD $\downarrow$ & FPVD $\downarrow$ & & JSD $\downarrow$ & \makecell{MMD $\downarrow$ \\ ($10^{-4}$)} \\
\midrule
Noise &  3277    & 497.1  & 336.2  & & 0.360   & 32.09 \\
LiDARGAN~\cite{caccia2019deep} & 1222    & 183.4  & 168.1 & & 0.272   & 4.74  \\
LiDARVAE~\cite{caccia2019deep}  & 199.1    & 129.9  & 105.8 & & 0.237   & 7.07   \\
ProjectedGAN~\cite{sauer2021projected} & 149.7   &   44.7   &   33.4   & &   0.188   &       2.88  \\
UltraLiDAR~\cite{xiong2023learning} & 370.0   &   72.1   &   66.6   & &  0.747   &       17.12  \\
LiDARGen~\cite{zyrianov2022learning} (1160s) &  129.1    & 39.2  & 33.4 & & \textbf{0.188}   & \textbf{2.88}  \\  % lidargen official 14k
\midrule
LiDARGen~\cite{zyrianov2022learning} (50s) & 2051    & 480.6  & 400.7 & & 0.506   & 9.91 \\
\rowcolor{baseline} LDM~\cite{rombach2022high} (50s)$^\dagger$ & 199.5    & 70.7  & 61.9  & & 0.236   & 5.06 \\
\rowcolor{ours} LiDM (ours, 50s)$^\dagger$ & 158.8     & 53.7    & 42.7   & & 0.213     & 4.46 \\ 
\rowcolor{ours} $\Delta$ \emph{Improv.} &  
{\scriptsize \color{PineGreen}{\bf{20.4\%}}} & 
{\scriptsize \color{PineGreen}{\bf{24.0\%}}} & 
{\scriptsize \color{PineGreen}{\bf{31.0\%}}} & & 
{\scriptsize \color{PineGreen}{\bf{9.7\%}}} &
{\scriptsize \color{PineGreen}{\bf{11.9\%}}}
\\
\rowcolor{ours_final} LiDM (ours, 50s) & \textbf{125.1} &   \textbf{38.8}   &  \textbf{29.0}   & &  0.211   &       3.84 \\ 
\rowcolor{ours_final} $\Delta$ \emph{Improv.} &  
{\scriptsize \color{PineGreen}{\bf{37.3\%}}} & 
{\scriptsize \color{PineGreen}{\bf{45.1\%}}} & 
{\scriptsize \color{PineGreen}{\bf{53.2\%}}} & & 
{\scriptsize \color{PineGreen}{\bf{10.6\%}}} &
{\scriptsize \color{PineGreen}{\bf{24.1\%}}}
\\
\bottomrule
\end{tabular}
\end{adjustbox}
\end{footnotesize}
\caption{
\label{tab:uncond}
Comparison of \textit{unconditional} LiDAR scene generation with recent state-of-the-art methods. We conduct experiments on 64-beam (\ie, KITTI-360~\cite{liao2022kitti}) data.``$\downarrow$'' indicates that lower values are better. $N$-s refers to $N$ sampling steps during inference. $\Delta$ \emph{Improv.} is the relative improvement of our method compared to the baseline of Latent Diffusion (LDM)~\cite{rombach2022high}, with the same number of diffusion steps. Note that, the diffusion process of LiDARGen~\cite{zyrianov2022learning} has 232 levels and 5 iterations in each level (\ie, 1160 steps in total). \crule[baseline]{7pt}{7pt} denotes baseline results, while \crule[ours]{7pt}{7pt} / \crule[ours_final]{7pt}{7pt} denotes our results. $\dagger$: Exactly the same settings except the architecture of autoencoders. We evaluate each method with 2000 randomly generated samples.
}
\vspace{-5mm}
\end{table}

%% file: table/cond_gen.tex
\begin{table*}[t]
\centering
\begin{footnotesize}
\begin{adjustbox}{max width=\linewidth}
\addtolength{\tabcolsep}{3pt}
\begin{tabular}{ccccccccccccccc}
\toprule
& \multicolumn{5}{c}{\textbf{Semantic-Map-to-LiDAR}~\cite{behley2019semantickitti}} & & \multicolumn{5}{c}{\textbf{Camera-to-LiDAR}~\cite{liao2022kitti}} \\
\cmidrule{2-6} \cmidrule{8-12}
\textbf{Method} & FRID $\downarrow$ & FSVD $\downarrow$ & FPVD $\downarrow$ & JSD $\downarrow$ & \makecell{MMD $\downarrow$ \\ ($10^{-4}$)} & & FRID $\downarrow$ & FSVD $\downarrow$ & FPVD $\downarrow$ & JSD $\downarrow$ & \makecell{MMD $\downarrow$ \\ ($10^{-4}$)} \\
\cmidrule{1-6} \cmidrule{8-12}
LiDARGen~\cite{zyrianov2022learning}  & 42.5 & 31.7 & 30.1 & 0.130 & 5.18 &  & - & - & - & - & -\\
\rowcolor{baseline} Latent Diffusion~\cite{rombach2022high} & 24.0 & 21.3 & 20.3 & 0.088 & 3.73 & & 50.2 & 35.9 & 26.5 & 0.256 & 3.80 \\
\rowcolor{ours} LiDAR Diffusion (ours) & \textbf{22.9} & \textbf{20.2} & \textbf{17.7} & \textbf{0.072} & \textbf{3.16} & & \textbf{44.9} & \textbf{32.5} & \textbf{25.8} & \textbf{0.205} & \textbf{3.69}  \\
\bottomrule
\end{tabular}
\end{adjustbox}
\end{footnotesize}
\caption{
\label{tab:cond}
Comparison of \textit{conditional} LiDAR scene generation with recent state-of-the-art methods. We conduct Semantic-Map-to-LiDAR experiments on SemanticKITTI~\citep{behley2019semantickitti} and Camera-to-LiDAR on KITTI-360~\cite{liao2022kitti}. ``$\downarrow$'' indicates that lower values are better. We implement Semantic-Map-to-LiDAR on LiDARGen through the concatenation operation. Camera-to-LiDAR on LiDARGen is not viable through concatenation, and hence we do not report results in this setting.
}
% \vspace{-5mm}
\end{table*}

%% file: table/efficiency.tex
\begin{table}[t]
\centering
\begin{adjustbox}{max width=\linewidth}
\addtolength{\tabcolsep}{0pt}
\begin{footnotesize}
\begin{tabular}{ccccc}
\toprule
Method & Diffusion Size & Throughput$\uparrow$ & Infer.Speed$\uparrow$  \\ 
\midrule
LiDARGen~\cite{zyrianov2022learning} & 64$\times$1024 & 0.015 & 17.5 \\
\rowcolor{ours} LiDM (ours) & 16$\times$128 & \textbf{1.603} & \textbf{80.2} \\
\bottomrule
\end{tabular}
\end{footnotesize}
\end{adjustbox}
\caption{
\label{tab:efficiency} 
Efficiency of LiDARGen and our LiDMs on the 64-beam scenario. We compute the number of generated samples per second of each model as throughput. \emph{Infer. Speed} is the number of inference passes (\ie, one pass representes a diffusion step) per second. We test both models in one NVIDIA RTX 3090 and adjust batch size to make full use of 24GB GPU memory.
}
\vspace{-1em}
\end{table}

%% file: section/conclusion.tex
\section{Limitations}

Even though LiDM establishes a new state of the art, its generated samples still have a visual gap relative to real-world LiDAR data. 
As shown in previous work~\cite{rombach2022high}, diffusion models should be powerful enough to capture the semantic information of input data, and therefore we argue that  autoencoders are the key to failures when recovering most details of scenes. Our contributions are a step towards LiDAR-realistic autoencoders, however further work is still required. %
For example, autoencoders may reconstruct blurry boundaries between objects and the background, which though imperceptible on range images, may lead to visually unreasonable objects in 3D space.

\section{Conclusion}

We propose LiDAR Diffusion Models (LiDMs), a \textit{general-conditioning} framework for LiDAR scene generation. 
With a focus on preserving curve-like patterns as well as scene-level and object-level geometry, we design an efficient latent space for DMs to achieve LiDAR-realistic generation
This design empowers our LiDMs to achieve competitive performance in unconditional generation and state of the art in conditional generation under 64-beam scenario, and enables the controllability of LiDMs with diverse conditions, including semantic maps, camera views, and text prompts.
To the best of our knowledge, ours is the first method to successfully introduce conditioning to LiDAR generation.
%

%% file: section/supp/main.tex
\appendix

\input{section/supp/method}

\input{section/supp/exp}

\input{section/supp/visual}

%% file: section/supp/method.tex
\section{Further Explanation of Concepts}
\subsection{Range Images and Point Clouds Conversion}
\label{sec:conversion}

In Sec.~\ref{sec:rep}, we introduced range images as the modality for both input and output within LiDAR Diffusion Models (LiDMs). Subsequently, in Sec.~\ref{sec:problem}, we provided a concise overview of the conversion process from range images to point clouds. This section extends our implementation discourse by delving into more comprehensive details.

The depth values are logarithmically scaled. To covert the pixel value $v$ back into depth value, we define:
\begin{equation}
\mathrm{depth}=2^{\omega \times v} - 1,
\end{equation}
where $\omega$ is a predefined scale factor. Given the normalized location $(a, b)$ of pixel $x$, where $a, b \in [0, 1]$,  we can compute its yaw and pitch through:
\begin{align}
&\mathrm{yaw}=(2a - 1) \times \pi, \\
&\mathrm{pitch}=(1-b) \times (\mathrm{fov}_{up} - \mathrm{fov}_{down}) + \mathrm{fov}_{down},
\end{align}
where $\mathrm{fov}_{up}$ and $\mathrm{fov}_{down}$ are specified based on the sensor settings of different datasets. Through the above computation, we can obtain the 3D coordinate $p$ of the pixel $x$. Likewise, we implement the conversion from a point cloud to a range image by performing the inverse calculation.
For the 32-beam scenario, $\mathrm{fov}_{up}=10^{\circ}$, $\mathrm{fov}_{down}=-30^{\circ}$, $\omega=5.53$. For the 64-beam scenario, $\mathrm{fov}_{up}=3^{\circ}$, $\mathrm{fov}_{down}=-25^{\circ}$, $\omega=5.84$. 

The transition from range images to point clouds is characterized by a lossless conversion. Conversely, when converting from point clouds to range images, occlusions commonly emerge. This occurrence is intricately tied to the resolution of range images. At a lower predefined resolution, multiple neighboring points tend to converge within a single pixel of a range image. In contrast, with a higher resolution, the incidence of missing pixels markedly rises, resulting in sparser range images. Hence, it is important to appropriately define resolutions in diverse scenarios to encompass more points with little geometric loss and to main a high density of range images. In the context of a 32-beam scenario, we set $H=32$ and $W=1024$, while in the 64-beam scenario, we set $H=64$ and $W=1024$.

\subsection{Statistical Evaluation Metrics}

In this paper, we adopt common statistical metrics, Jensen-Shannon Divergence (JSD) and Minimum Matching Distance (MMD), for evaluation introduced in \cite{achlioptas2018learning} and adopted by some recent works~\cite{zyrianov2022learning, yang2019pointflow}. 

\textbf{Jensen-Shannon Divergence (JSD)} measures the degree to which point clouds of synthesized set $S$ tend to occupy the similar locations as those of reference set $R$. It can be defined as follows:
\begin{equation}
\mathrm{JSD}\left(P_S \| P_R\right)=\frac{1}{2} D_{KL}\left(P_R \| M\right)+\frac{1}{2} D_{KL}\left(P_S \| M\right),
\end{equation}
where $M=\frac{1}{2}(P_R + P_S)$ and $D_{KL}$ is KL divergence~\cite{kullback1951information}. In this paper, we compute JSD after discretizing each LiDAR point cloud into $2000^2$ voxels in the form of Birds' Eye View (BEV), with width and length of each voxel $0.05$.

\textbf{Minimum Matching Distance (MMD)} matches each LiDAR point cloud of reference set $R$ to the one in synthesized set $S$ with minimum distance and averages all distances in the matching. It indicates the fidelity of $S$ with respect to $R$. We define MMD as follows:
\begin{equation}
\mathrm{MMD}\left(P_S \| P_R\right)=\frac{1}{\left|P_R\right|} \sum_{Y \in P_R} \min _{X \in P_S} D_{CD}(X, Y).
\end{equation}
Considering efficiency, we choose Chamfer Distance (CD) instead of Earth Mover's Distance (EMD) to represent the distance of two LiDAR point clouds. Both are defined in Sec.~\ref{sec:dist}. Different from JSD, the computation of MMD requires traversing all reference samples for each synthesized sample, which results in larger amounts of computation. To guarantee its efficiency, we adopt a larger voxel size of $0.5$ to voxelize each point cloud into $200^2$ BEV.

\subsection{Perceptual Evaluation Metrics}

\subsubsection{Background}

In general, distinguished from statistical evaluation metrics, perceptual metrics describe the performance of generative models through a perceptual space provided by pretrained models. In light of the incompatibility of classification-based models in the context of LiDAR scenes, we opt for segmentation-based pretrained models to delineate the perceptual metrics proposed in this paper.

Similar to the widely adopted perceptual metrics Fréchet Image Distance (FID)~\cite{heusel2017gans} and Inception Score (IS)~\cite{szegedy2016rethinking} in image synthesis, we compute the results of our proposed perceptual metrics in the final stage. Given a trained UNet-like model $\Theta$ consisting of an encoder $\Theta_\mathcal{E}$ with $L$ layers and a decoder $\Theta_\mathcal{D}$ with $L$ layers, the output activation of a pixel (before dropout) from the final stage can be defined as:

\begin{equation}
a_{final} = \Theta^L_\mathcal{D}([x, \Theta^1_\mathcal{E}(x)]),
\end{equation}
where $a_{final} \in \mathbb{R}^{H \times W \times C}$.

%%%%%%%%%%%%%%%%%%%%%%%%%%%%%%%%%%%%%%%%%%%%%%%%%%%%%%%%%%%%%
\begin{figure}
\begin{center}
\includegraphics[width=0.48\textwidth]{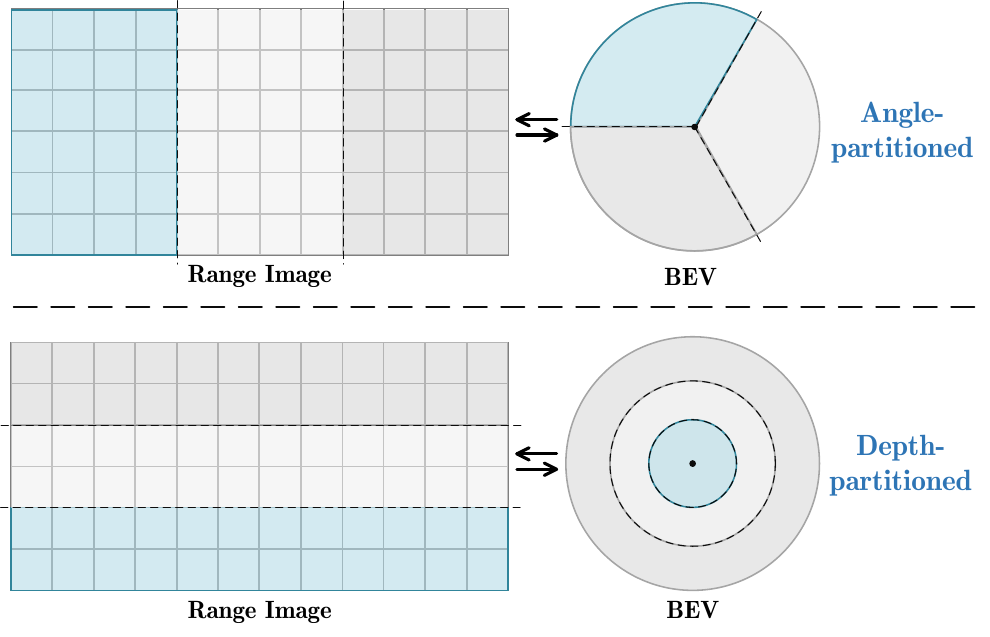}
\end{center}
\vspace{-0.2in}
\caption{Example for two manners of partition-based aggregation on range images and bird's eye view (BEV). Angle-partitioned aggregation performs average on partitions of several columns on range images and of a sector on BEVs, while depth-partitioned aggregation performs average on partitions of several rows on range images and of a ring on BEVs. In this paper, we adopt depth-partitioned aggregation by default for its rolling-operation-invariant ability.}
\label{fig:agg} 
\end{figure}
%%%%%%%%%%%%%%%%%%%%%%%%%%%%%%%%%%%%%%%%%%%%%%%%%%%%%%%%%%%%%

\subsubsection{Aggregation Manners} 

Unlike classification-based network, the output of segmentation-based networks is a map of activations. Therefore, we cannot directly obtain the global feature of the input. In this paper, we provide two possible manners, \textit{angle}-partitioned aggregation and \textit{depth}-partitioned aggregation, to approximately represent the global feature given the output map of activations of an input range image. An illustration is shown in Fig.~\ref{fig:agg}. To obtain the global feature of one LiDAR point cloud, we uniformly divide it into $P$ parts and concatenate all of them after average pooling on each part, resulting in a vector with $P \times C$ channels.

As shown in Fig.~\ref{fig:agg} \textit{Above}, angle-partitioned aggregation partitions each LiDAR point cloud into $P$ sectors by yaw angle. Since $x$-coordinate of each pixel on a range image is defined through linear transformation of the yaw angle of a point (\cf, Sec.~\ref{sec:conversion}), the range image is partitioned into $P$ regions, and each is represented by $W/P$ columns. Each sector has an equivalent region in the range image.

Similarly, in Fig.~\ref{fig:agg} \textit{Below}, depth-partitioned aggregation splits a point cloud into $P$ rings in the BEV level and $P$ regions represented by $H/P$ rows in the range-image level. Note that, different from angle-partitioned aggregation, the divided ring of the point cloud and its corresponding region of the range image are not equivalent in each pair. 

Since the LiDAR point clouds are density-varying with depth, depth-partitioned aggregation is density-aware. Contrarily, the partitions by angle ignore the depth and each represent a sub-LiDAR-point-cloud. In this paper, we default to the utilization of depth-partitioned aggregation, as it effectively avoids the variability from rolling operation associated with angle-partitioned aggregation.

\subsubsection{Implementation Details} 

In this paper, we propose three perceptual metrics: Fréchet Range Image Distance (FRID), Fréchet Sparse Volume Distance (FSVD), and Fréchet Point-based Volume Distance (FPVD) and set the number of partitions $P=16$ by default. For each proposed perceptual metric, we further provide its details as follows:
\begin{itemize}
\item \textbf{FRID:} RangeNet++~\cite{milioto2019rangenet++} is a range-image-based method to predict per-pixel semantic labels. It adopts various image-based UNet for training. In this paper, we adopt a DarkNet21-based~\cite{redmon2013darknet} model trained through the official implementation\footnote{\url{https://github.com/PRBonn/lidar-bonnetal}}. With the trained model, we can easily obtain the output in the final stage, which is in the shape of $64 \times 1024 \times 32$. We derive a global feature vector of each range image with $512$ channels followed by a spatial averaging pooling. It is noteworthy that our proposed FRID effectively addresses the issue of result instability arising from random sampling, as indicated by the FRD score introduced in \cite{zyrianov2022learning}.
\item \textbf{FSVD:} Sparse volumes are a prevalent 3D modality in LiDAR scenes. Unlike range images, volumes can directly represent 3D shapes without projection. To compute FSVD, we adopt a simple backbone, MinkowskiNet~\cite{choy20194d}, to extract features from the sparse volumes converted from range images. We utilize a public implementation with the pretrained weights\footnote{\label{note1}Implementation from \url{https://github.com/yanx27/2DPASS}} based on torchsparse~\cite{tang2022torchsparse}. We calculate the average of all active (\ie, non-empty) voxel features for each partition in the final stage, resulting in a $1536$-channel vector.
\item \textbf{FPVD:} Leveraging the support of point clouds, the hybrid of point clouds and sparse volumes preserves a richer set of geometric information compared to utilizing sparse volumes alone. In the calculation of FPVD, we employ SPVCNN~\cite{tang2020searching} as the backbone, utilizing the public implementation as in FSVD. The computational process of FPVD is the same as FSVD, with the output of $1536$-channel global features.
\end{itemize}

\subsection{Details of Training}

\subsubsection{Perceptual Loss for LiDAR Compression}

%%%%%%%%%%%%%%%%%%%%%%%%%%%%%%%%%%%%%%%%%%%%%%%%%%%%%%%%%%%%%
\begin{figure}
\begin{subfigure}{.23\textwidth}
\centering
\includegraphics[width=\linewidth]{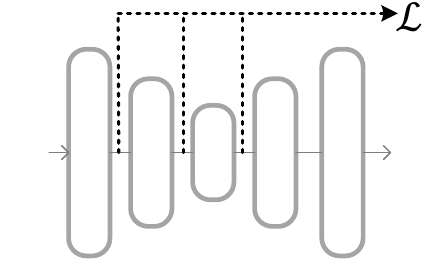}
\caption{Supervision on encoder}
\end{subfigure}
\begin{subfigure}{.23\textwidth}
\centering
\includegraphics[width=\linewidth]{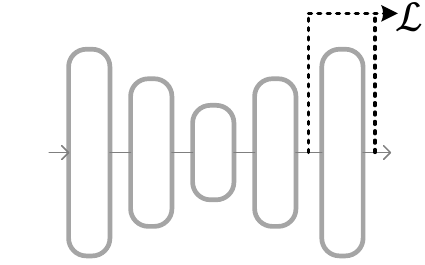}
\caption{Supervision on decoder}
\end{subfigure}
\begin{subfigure}{.23\textwidth}
\centering
\includegraphics[width=\linewidth]{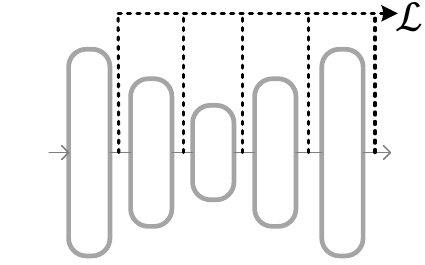}
\caption{Supervision on all layers}
\end{subfigure}
\begin{subfigure}{.23\textwidth}
\centering
\includegraphics[width=\linewidth]{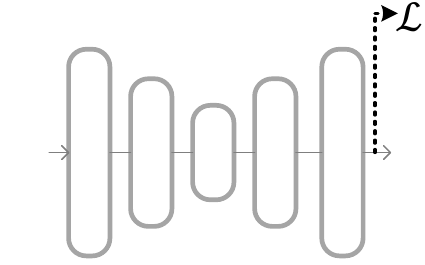}
\caption{Supervision on the final layer}
\end{subfigure}
\caption{Different types of feature extraction for the computation of perceptual loss.}
\label{fig:perceptual}
\vspace{-0.3cm}
\end{figure}
%%%%%%%%%%%%%%%%%%%%%%%%%%%%%%%%%%%%%%%%%%%%%%%%%%%%%%%%%%%%%

The regularization of models based on the pixel-wise depth of the synthesized range image and the ground-truth range image has the potential to disproportionately penalize outputs that are, in fact, LiDAR-realistic. For instance, given the same 2D projected shapes and scales in range images, generating cars farther away from the ego-center than closer to the ego-center induces a high loss. To mitigate this challenge, we leverage the success demonstrated by perceptual loss in the domain of image synthesis.

\cite{gatys2016image} introduced the output of different image processing stages in a pretrained VGG network~\cite{simonyan2014very} as the ``content representation''. This idea was subsequently evolved as a perceptual regularization to learn both fine-grained details by matching lower-layer activations and global part arrangement by matching higher-layer activations. This regularization is widely adopted in various image tasks, including image super-resolution~\cite{johnson2016perceptual, bruna2015super}, neural style transfer~\cite{gatys2016image}, and image synthesis~\cite{dosovitskiy2016generating, chen2017photographic}. Some recent popular generators~\cite{rombach2022high, esser2021taming} are trained in a perceptual space based on LPIPS~\cite{zhang2018unreasonable}, a common learned perceptual metric to evaluate image synthesis performance. Unfortunately, this ``learned'' metric is not available in the context of LiDAR scenes, and thus in this paper, we design perceptual loss by matching the output activations of stages with different scales, similar to the perceptual loss introduced in \cite{chen2017photographic}.

We utilize pretrained segmentation-based networks (\eg, RangeNet++~\cite{milioto2019rangenet++}) to extract features as the preparation of feature matching. We explore four variants of perceptual loss based on different types of feature extraction. An illustration is shown in Fig.~\ref{fig:perceptual}.

\subsubsection{Mask Prediction Loss in LiDAR Compression}

Though range images are dense representation for LiDAR scans, they contain a large number of invalid pixels. To distinguish them from valid pixels, our autoencoder outputs a binary mask along with the resulting range image. To this end, we apply mask prediction loss to our reconstruction loss and adversarial loss as follows:
\begin{align}
&\mathcal{L}_{rec}\left(x\right)=\mathbb{E}_x[\|x-\hat{x}\| + \lambda_1\|p-\hat{p}\|_2^2 + \lambda_2\|m-\hat{m}\|_2^2], \\
&\mathcal{L}_{\mathrm{GAN}}(x)=\mathbb{E}_x[\log \mathcal{D}([x, p, m])+\log (1-\mathcal{D}([\hat{x}, \hat{p}, \hat{m}]))],
\end{align}
where $m$ is the mask value corresponding to $x$.

%% file: section/supp/exp.tex
\section{Additional Experimental Results}

\input{table/supp/autoencoder}

\input{table/supp/lidm}

\subsection{Design of Autoencoders for LiDAR Compression}
\label{sec:autoencoder}

\subsubsection{Settings}
\label{sec:dist}

To study on the behavior of LiDAR compression with different manners and ratios of downsampling, we provide our comprehensive studies on the design of autoencoders for LiDAR compression. In Table~\ref{tab:autoencoder}, we conduct experiments on various downsampling factors $f_c$ and $f_p$ for curve-wise and/or patch-wise encoding, respectively. To set up a comparable test field, we fix computational resources to four NVIDIA RTX 3090 GPUs and training steps to 40k steps for all listed experiments. We train autoencoders on KITTI-360~\cite{liao2022kitti} and evaluate the quality of reconstruction in autoencoding process with perceptual metrics of reconstruction, \ie, R-FRID, R-FSVD, R-FPVD, and statistical metrics, \ie, Chamfer Distance (CD), Earth Mover's Distance (EMD). Following prior works~\cite{achlioptas2018learning, yang2019pointflow}, we define CD and EMD as follows:
\begin{align}
&\mathrm{CD}(P_x, P_{\hat{x}}) =\sum_{p \in P_x} \min _{\hat{p} \in P_{\hat{x}}}\|p-\hat{p}\|_2^2+\sum_{\hat{p} \in P_{\hat{x}}} \min _{p \in P_x}\|p-\hat{p}\|_2^2, \\
&\mathrm{EMD}(P_x, P_{\hat{x}}) =\min _{\phi: P_x \rightarrow P_{\hat{x}}} \sum_{p \in P_x}\|p-\phi(p)\|_2,
\end{align}
where $P_x$ and $P_{\hat{x}}$ are the ground-truth and reconstructed point clouds and $\phi$ is a bijection between them. Note that, $P_x$ and $P_{\hat{x}}$ are the point clouds projected back from range images $x$ and $\hat{x}$ instead of the raw point clouds.

\subsubsection{Analysis and Discussion}

As shown in \cite{rombach2022high}, performance of image compression is highly related with the synthesis quality of DMs. By analyzing the results of either curve-wise or patch-wise encoding in Table~\ref{tab:autoencoder}, we conclude several valuable clues for the design of autoencoders: for curve-wise encoding (\textbf{Curve}), (i) as indicated by metrics R-FSVD and R-FPVD, the quality in the point-cloud level decreases with $f_c$ increasing, and (ii) when $f_c \in \{4,8,16\}$, curve-wise encoding strikes better perceptually faithful results, while for patch-wise encoding (\textbf{Patch}), (i) when $f_p=4$, with the same overall scale factor $f$, patch-wise encoding results in comparable reconstructed results of curve-wise encoding with $f_c=16$, and (ii) when $f=4$, curve-wise encoding outperforms patch-wise encoding by a large margin in both point-cloud and range-image level.

Curve-wise and patch-wise encoding can be complementary: curve-wise encoding learns within horizontal receptive fields to capture the curve-like structures existing in range images, and patch-wise encoding after curve-wise encoding vertically extends the receptive fields to learn object-level information. Following this nature, we design autoencoders to compress range images through both curve-wise and patch-wise encoding as hybrid encoding. 

Based on the aforementioned analysis, we conduct studies on hybrid encoding (\textbf{Hybrid}) with diverse settings, keeping overall scale $8 \leq f \leq 64$. The results are listed in Table~\ref{tab:autoencoder}. Considering both performance and efficiency (overall scale), we select two settings: (a) $f_c=2$, $f_p=2$, and (b) $f_c=2$, $f_p=4$. We indicate that Model (a) outperforms all settings of curve-wise and patch-wise encoding when $f \geq 8$, and Model (b) achieves both competitive performance and high compression rate.

\subsection{Design of LiDAR Diffusion Models}

\subsubsection{Settings}

In Sec.~\ref{sec:autoencoder}, we report the performance of autoencoders with different encoding manners and scaling factors. Although the reconstruction performance on validation set and the synthesis quality of DMs show a strong relation, the two are not always \textit{positively} correlated. Thus, to further explore the behavior of LiDAR compression, we conduct experiments to train DMs with the trained autoencoders. The results are reported in Table~\ref{tab:lidm}. We train each DM with 10k steps and adopt the same experimental setup in Sec.~\ref{sec:autoencoder}. We apply 50 sampling DDIM~\cite{song2020denoising} steps to each model and generate 5,000 samples for evaluation.

\subsubsection{Analysis and Discussion}

Through the experimental results in Table~\ref{tab:lidm}, we conclude a similar fact as in Sec.~\ref{sec:autoencoder} that hybrid encoding generally performs much better than curve-wise or patch-wise encoding with the same compression rate. However, though the reconstruction performance of Model (a) ($f_c=2$, $f_p=2$)  performs the best among all settings in Table~\ref{tab:autoencoder}, Model (b) ($f_c=2$, $f_p=4$) generates samples with better synthesis quality under an attractive compression rate. Therefore, in this paper, we adopt Model (b) ($f_c=2$, $f_p=4$) as the default autoencoder for LiDAR compression.

%% file: table/supp/autoencoder.tex
\begin{table*}[t]
\centering
\begin{footnotesize}
\begin{adjustbox}{max width=\linewidth}
\addtolength{\tabcolsep}{0pt}
\begin{tabular}{ccccc|cl|cccccc}
\toprule
& $f_c$ & $f_p$ & $c$ & $|\mathcal{Z}|$ & \makecell{Overall Scale $f$} & Encoded Size &  R-FRID $\downarrow$ & R-FSVD $\downarrow$ & R-FPVD $\downarrow$ & CD $\downarrow$ & EMD $\downarrow$ & \#Params (M) \\
\midrule
\multirow{5}{*}{\textbf{Curve}} 
& 4 & 1 & 2 & 4096 &  4 & $64 \times 256 \times 2$     & \textbf{0.2}     &     \textbf{12.9}      &     \textbf{13.8}      & \textbf{0.069} & \textbf{0.151} & 9.52 \\
& 8 & 1 & 3 & 8192 &  8  & $64 \times 128 \times 3$    & \underline{0.9}     &     \underline{21.2}      &     \underline{17.4}      & \underline{0.141} & \underline{0.230} & 10.76 \\
& 16 & 1 & 4 & 16384 &  16 & $64 \times 64 \times 4$   & 2.8     &     31.1      &     23.9      & 0.220 & 0.265 & 12.43 \\ 
& 32 & 1 & 8 & 16384 &  32  & $64 \times 32 \times 8$  & 16.4     &     49.0      &     38.5      & 0.438 & 0.344 & 13.72 \\ 
& 64 & 1 & 16 & 16384 &  64 & $64 \times 16 \times 16$ & 34.1     &     98.4      &     83.7      & 0.796 & 0.437 & 20.06 \\ 
\midrule
\multirow{4}{*}{\textbf{Patch}} 
& 1 & 2 & 2 & 4096 &  4 & $32 \times 512 \times 2$     & \underline{1.5}     &     \underline{25.0}      &     \underline{23.8}      & \textbf{0.096} & \textbf{0.178} & 2.87 \\
& 1 & 4 & 4 & 16384 &  16 & $16 \times 256 \times 4$   & \textbf{0.6}     &     \textbf{15.4}      &     \textbf{15.8}      & \underline{0.142} & \underline{0.233} & 12.45 \\
& 1 & 8 & 16 & 16384 &  64 & $8 \times 128 \times 16$  & 17.7     &     35.7      &     33.1      & 0.384 & 0.327 & 15.78 \\
& 1 & 16 & 64 & 16384 &  256 & $4 \times 64 \times 64$ & 37.1     &     68.7      &     63.9      & 0.699 & 0.416 & 16.25 \\
\midrule
 \multirow{6}{*}{\makecell{\textbf{Hybrid} \\ ($8 \leq f \leq 64$)}} 
& \hp 2 & \hp 2 & \hp 3 & \hp 8192 &  \hp 8 & \hp $32 \times 256 \times 3$     & \hp \textbf{0.4}     &    \hp \textbf{11.2}      &    \hp \textbf{12.2}      & \hp \textbf{0.094} & \hp \textbf{0.199} & \hp 13.09 \\
& 4 & 2 & 4 & 16384 &  16 & $32 \times 128 \times 4$   & 3.9     &     19.6      &     16.6      & \underline{0.197} & \underline{0.236} & 14.35 \\
& 8 & 2 & 8 & 16384 &  32 & $32 \times 64 \times 8$    & 8.0     &     25.3      &     20.2      & 0.277 & 0.294 & 16.06 \\ 
& 16 & 2 & 16 & 16384 &  64 & $32 \times 32 \times 16$ & 21.5     &     54.2      &     44.6      & 0.491 & 0.371 & 17.44 \\
\cmidrule{2-13}
 & \ours 2 & \ours 4 & \ours 8 & \ours 16384 &  \ours 32 & \ours $16 \times 128 \times 8$   & \ours \underline{2.5}     &    \ours \underline{16.9}      &    \ours \underline{15.8}      & \ours 0.205 & \ours 0.273 & \ours 15.07 \\
& 4 & 4 & 16 & 16384 &  64 & $16 \times 64 \times 16$ & 13.8     &     29.5      &     25.4      & 0.341 & 0.317 & 16.86 \\
\bottomrule
\end{tabular}
\end{adjustbox}
\end{footnotesize}
\caption{
\label{tab:autoencoder}
Performance of autoencoders in different downsampling factors $f_c$ and $f_p$ \textit{after 40k training steps} on the KITTI-360 \textit{val}~\cite{liao2022kitti}. $f_c$ is the curve-wise encoding factor, and $f_p$ is the patch-wise encoding factor. $f=f_c \times f_p^2$ is the overall scaling factor. Encoded size ($h \times w \times c$) is the output after encoding, where $h = H/f_p$ and $w= W/(f_c \times f_p)$. We evaluate the reconstruction quality of the trained autoencoders through reconstruction-based perceptual metrics (\ie, R-FRID, R-FSVD, R-FPVD) and statistical pairwise metrics (\ie, CD, EMD). For comparison of \textit{each} encoding manner, \textbf{bold} means the best in one metric, and \underline{underline} means the second best.
}
\end{table*}

%% file: table/supp/lidm.tex
\begin{table*}[t]
\centering
\begin{footnotesize}
\begin{adjustbox}{max width=\linewidth}
\addtolength{\tabcolsep}{0pt}
\begin{tabular}{ccccc|cl|cccccc}
\toprule
& $f_c$ & $f_p$ & $c$ & $|\mathcal{Z}|$ & \makecell{Overall Scale $f$} & Encoded Size &  FRID $\downarrow$ & FSVD $\downarrow$ & FPVD $\downarrow$ & JSD $\downarrow$ & MMD ($\times 10^{-4}$) $\downarrow$ & \#Params (M) \\
\midrule
\multirow{3}{*}{\textbf{Curve}} 
& 4 & 1 & 2 & 4096 &  4 & $64 \times 256 \times 2$ & 271 & 148 & 118 & 0.262 & 5.33 & \textcolor{blue}{9.5}$+$\textcolor{red}{36$^*$} \\
& 8 & 1 & 3 & 8192 &  8  & $64 \times 128 \times 3$ & 162 & \textbf{85} & \textbf{68} & 0.234 & \textbf{5.03} & \textcolor{blue}{10.8}$+$\textcolor{red}{258} \\
& 16 & 1 & 4 & 16384 &  16 & $64 \times 64 \times 4$ & \textbf{142} & 116 & 106 & \textbf{0.232} & 5.15  & \textcolor{blue}{11.1}$+$\textcolor{red}{258} \\ 
\midrule
\multirow{3}{*}{\textbf{Patch}} 
& 1 & 2 & 2 & 4096 &  4 & $32 \times 512 \times 2$  &  205 & 154 & 132 & 0.248 & 6.15  & \textcolor{blue}{2.9}$+$\textcolor{red}{36$^*$}  \\
& 1 & 4 & 4 & 16384 &  16 & $16 \times 256 \times 4$ & \textbf{180} & \textbf{60} & \textbf{55} & \textbf{0.230} & \textbf{5.34} & \textcolor{blue}{12.5}$+$\textcolor{red}{258}    \\
& 1 & 8 & 16 & 16384 &  64 & $8 \times 128 \times 16$  & 192 & 88 & 78 & 0.243 & 5.14 &   \textcolor{blue}{15.8}$+$\textcolor{red}{258} \\
\midrule
 \multirow{6}{*}{\makecell{\textbf{Hybrid} \\ ($8 \leq f \leq 64$)}} 
& \hp 2 & \hp 2 & \hp 3 & \hp 8192 &  \hp 8 & \hp $32 \times 256 \times 3$   & \hp 161 & \hp 73 & \hp 63 & \hp 0.228 & \hp 5.44 & \hp \textcolor{blue}{13.1}$+$\textcolor{red}{258}  \\
& \hp 2 & \hp 2 & \hp 3 & \hp 16384 &  \hp 8 & \hp $32 \times 256 \times 3$   & \hp 165 & \hp 76 & \hp 65 & \hp 0.231 & \hp 5.28 & \hp \textcolor{blue}{13.1}$+$\textcolor{red}{258}  \\
& 4 & 2 & 4 & 16384 &  16 & $32 \times 128 \times 4$ & \textbf{145} & 77 & 68 & \textbf{0.222} & 5.10 & \textcolor{blue}{14.4}$+$\textcolor{red}{258} \\
& 8 & 2 & 8 & 16384 &  32 & $32 \times 64 \times 8$  & 188 & 83 & 71 & 0.228 & 5.33 & \textcolor{blue}{16.1}$+$\textcolor{red}{258}  \\ 
\cmidrule{2-13}
& \ours 2 & \ours 4 & \ours 8 & \ours 16384 &  \ours 32 & \ours $16 \times 128 \times 8$ & \ours 162 & \ours \textbf{56} & \ours \textbf{49} & \ours 0.228 & \ours \textbf{4.82} & \ours \textcolor{blue}{15.1}$+$\textcolor{red}{258}  \\
& 4 & 4 & 16 & 16384 &  64 & $16 \times 64 \times 16$ & 195 & 80 & 70 & 0.240 & 5.84 &     \textcolor{blue}{16.9}$+$\textcolor{red}{258} \\
\bottomrule
\end{tabular}
\end{adjustbox}
\end{footnotesize}
\caption{
\label{tab:lidm}
Performance of LiDMs with autoencoders in different downsampling factors $f_c$ and $f_p$ \textit{after 10k training steps} on the KITTI-360 \textit{val}~\cite{liao2022kitti}. $f_c$ is the curve-wise encoding factor, and $f_p$ is the patch-wise encoding factor. $f=f_c \times f_p^2$ is the overall scaling factor. Encoded size ($h \times w \times c$) is the output after encoding, where $h = H/f_p$ and $w= W/(f_c \times f_p)$. We evaluate the synthesis quality of the trained LiDMs through perceptual metrics (\ie, FRID, FSVD, FPVD) and statistical metrics (\ie, JSD, MMD). For comparison of \textit{each} encoding manner, \textbf{bold} means the best in one metric. We present the number of parameters (\#Params) with \textcolor{blue}{blue} for the autoencoder part and \textcolor{red}{red} is for the diffusion model part. $*$: Modification on the number of basic channels for appropriate GPU memory cost.
}
\end{table*}

%% file: section/supp/visual.tex
\section{Additional Qualitative Results}

\subsection{32-Beam Unconditional LiDAR Generation}

\begin{figure*}
\begin{center}
\includegraphics[width=0.75\textwidth]{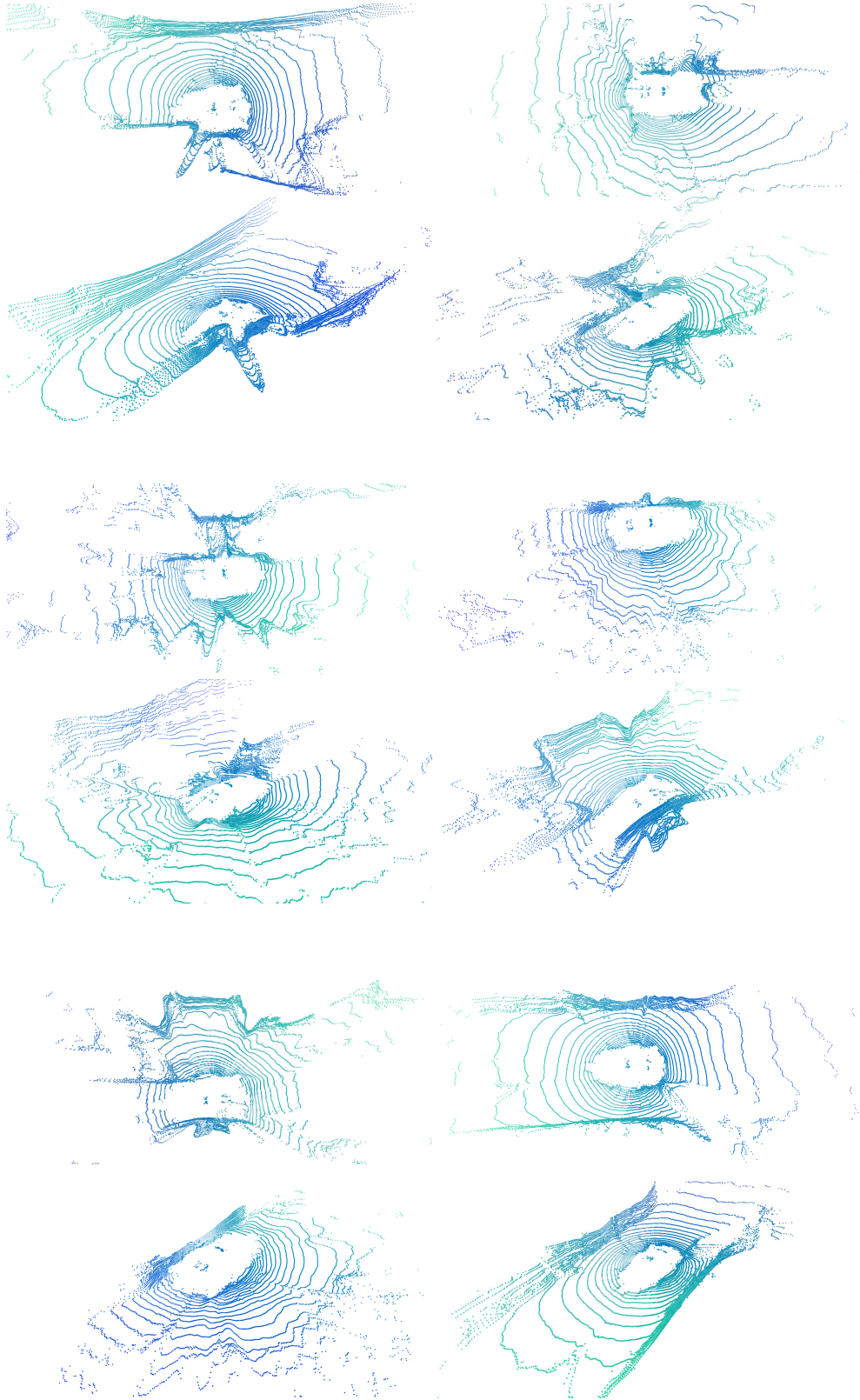}
\end{center}
   \caption{Unconditional samples on 32-beam scenario.}
\label{fig:uncond32} 
\end{figure*}

In Fig.~\ref{fig:uncond32}, we visualize the results unconditional LiDM on 32-beam data. 32-beam results appear sparser and noisier than the 64-beam results. We argue that this is highly related to the density and quality of the collected LiDAR point clouds. The 64-beam dataset, KITTI-360~\cite{liao2022kitti}, provides point clouds with denser foreground objects and clear boundaries between objects and backgrounds (\eg, walls, roads). LiDMs benefit from the dense data, and thus can recognize objects more easily and learn from the geometry of complex 3D scenes.

